\documentclass{article}

\PassOptionsToPackage{numbers, compress}{natbib}

\usepackage{url}
\usepackage{hyperref}

\usepackage{listings}
\usepackage{xcolor}
\usepackage{textcomp}
\definecolor{codegreen}{rgb}{0,0.6,0}

\definecolor{backcolour}{rgb}{0.95, 0.95, 0.92}

\usepackage{textcomp}
\usepackage[preprint]{corl_2026} 
\usepackage{graphicx}
\usepackage{amsmath}

\usepackage{caption}
\usepackage{subcaption}
\usepackage{array}
\usepackage{tabularx}

\usepackage{etoc}
\usepackage{minitoc}
\usepackage{bbm}

\usepackage{url}
\usepackage[capitalize,noabbrev]{cleveref}
\usepackage{wrapfig}
\crefname{section}{Section}{Sections}
\Crefname{section}{Section}{Sections}
\Crefname{table}{Table}{Tables}
\crefname{table}{Table}{Tables}
\Crefname{figure}{Figure}{Figures}
\crefname{figure}{Figure}{Figures}
\crefname{subfigure}{Figure}{Figures}
\Crefname{subfigure}{Figure}{Figures}

\usepackage{booktabs}
\usepackage{multirow}
\usepackage{xcolor}

\title{PhysCaP\hspace{0.05cm}\raisebox{-.15\height}{\includegraphics[scale=0.02]{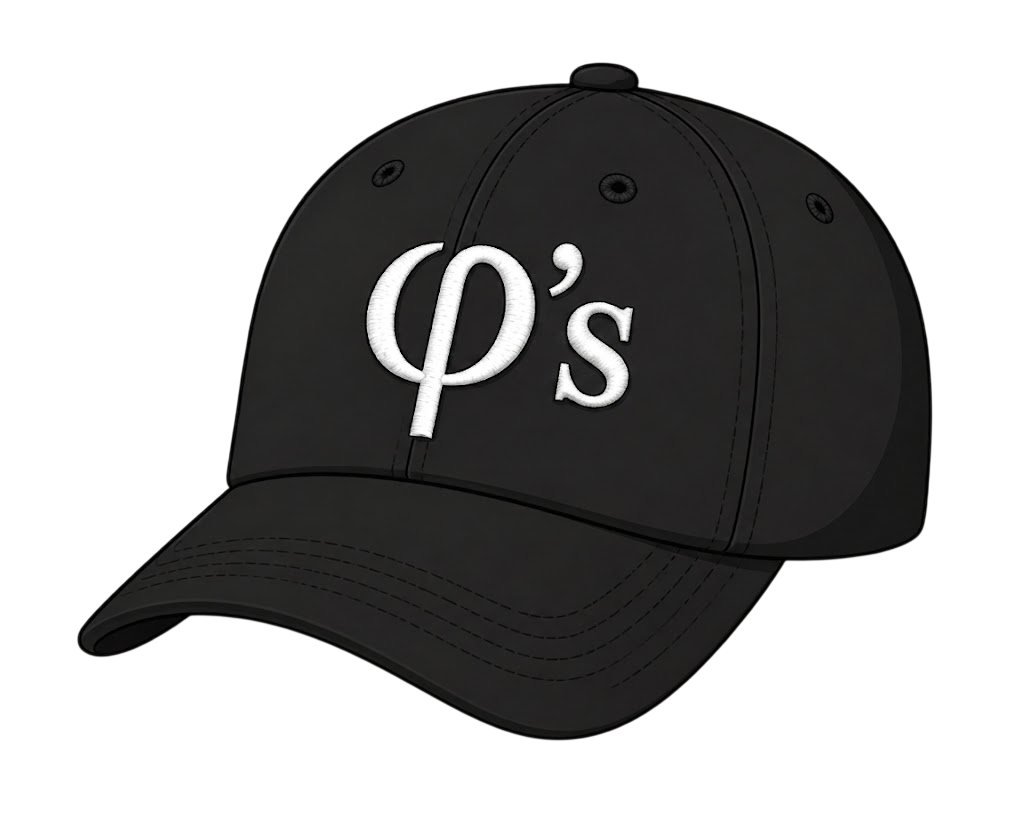}}: Grounding Code-as-Policy Agent with Physics-Informed Exploration}

\begingroup
\hypersetup{colorlinks=false, linkcolor=black}
\hypersetup{pdfborder={0 0 0}}

\author{
  Chen-Yu Lin$^{1*}$ \quad Jing-Wen Chen$^{1*}$ \quad Hsueh-En Chang$^1$ \quad Hung-An Chen$^1$\\
  \textbf{Sheng-Hsun Chang$^1$  \quad 
  Chi-Pin Huang$^2$ \quad Fu-En Yang$^2$ \quad Min-Hung Chen$^2$}\\
 \textbf{Yi-Ting Chen$^3$ \quad Yu-Chiang Frank Wang$^2$ \quad Shao-Hua Sun$^1$}\\
  $^1$National Taiwan University \quad
  $^2$NVIDIA Research \quad 
  $^3$National Yang Ming Chiao Tung University}

\endgroup

\usepackage{xcolor}         

\newcommand{\eg}{\textit{e}.\textit{g}.,\ }

\newcommand{\method}{PhysCaP}

\newcommand{\vspacesection}[1]{\vspace{-0.13cm}
\section{#1}
\vspace{-0.13cm}}
\newcommand{\vspacesubsection}[1]{\vspace{-0.13cm}
\subsection{#1}
\vspace{-0.13cm}}

\newcommand\blfootnote[1]{%
  \begingroup
  \begin{NoHyper}%
  \renewcommand\thefootnote{}\footnote{#1}%
  \addtocounter{footnote}{-1}%
  \end{NoHyper}%
  \endgroup
}

\begin{document}

\maketitle


\begin{abstract}
We present PhysCaP, a Physics-Informed Code-as-Policy agent system for active perception in robotic manipulation. While vision-language-action policies excel at imitating demonstrations, they rely on passive observation and fail to infer latent physical properties critical for manipulation. PhysCaP augments code-as-policy frameworks with a physics-informed exploration layer that enables explicit information-seeking through interaction. Our method introduces training-free physical property extraction modules that estimate object mass and stiffness from robot proprioception without additional sensors. To balance exploration costs and the efficiency of information obtained, PhysCaP employs a multi-agent design: a Planner that decides when to explore and when to stop, and a Prioritizer that filters implausible interactions and ranks the remainder using a heuristic priority score, enabling efficient, targeted exploration. We evaluate PhysCaP on three real-world tabletop manipulation tasks and a simulated task in LIBERO. The results show that existing passive and naive interactive baselines either fail when physical properties are hidden or over-explore, whereas PhysCaP achieves comparable performance with fewer interactions and reduced execution time. Ablation studies further validate the effectiveness of the proposed physical property extraction modules. \\Project page: \href{https://physcap.github.io}{https://physcap.github.io}
\blfootnote{$^*$Equal contribution}
\end{abstract}



\vspacesection{Introduction}
\label{sec:intro}

Actively acquiring information through physical interaction is a core aspect of human intelligence \cite{gibson1979ecological, bajcsy1988active}. 
Rather than relying solely on passive observation, humans actively interact with the environment to reduce uncertainty, for example by uncovering occluded objects, estimating physical properties, or predicting object dynamics. When cleaning a table, for instance, a person may lift or shake cans to infer whether they are empty, using prior knowledge and common sense, \eg unopened cans are unlikely to be empty, to avoid unnecessary interactions.

Recent vision-language-action (VLA) policies \cite{zitkovich2023rt, team2024octo, kim2024openvla, intelligence2025pi_, huang2025thinkact, fang2026molmoact2} trained with large-scale imitation learning achieve strong performance on complex manipulation tasks such as cloth folding and object rearrangement. Such policies can, in principle, exhibit latent information-seeking behavior \cite{huang2025thinkact} when it is represented in their training distribution or learned through interaction. However, acquiring task-specific exploratory behaviors through imitation or reinforcement learning \cite{kerr2025eyerobot, hu2025aawr} can require substantial data, while the resulting sensing strategy is often implicit in the learned policy \cite{aoyama2025poke} and difficult to inspect or modify. This becomes particularly challenging when a new task requires a physical quantity that is not directly available from the policy's existing observations.

These limitations motivate an agentic formulation of active perception, where information seeking becomes an explicit part of decision-making. Such agents can identify task-relevant uncertainty \cite{sun2024interactive, nazarczuk2025closed}, invoke task-dependent interaction tools to resolve it \cite{zhao2023chat, jiang2024roboexp}, and incorporate the resulting evidence into subsequent decisions. Code-as-policy systems \cite{ahn2022can, huang2022innermonologue, song2023llmplanner, wang2023voyager, wang2024codeact, wang2025vlmseerobotdo} are particularly well suited to this setting because executable programs can compose perception \cite{clark2026molmo2} and control APIs \cite{team2024gemini} while incorporating new physical sensing capabilities without retraining the underlying model. However, this flexibility does not by itself provide access to latent physical properties: an agent may recognize that mass or stiffness is relevant yet lack a mechanism to measure them. Once such measurement capabilities are available, the agent must also balance the risk of acting with insufficient information against the cost of unnecessary interactions.

\begin{wrapfigure}{r}{0.5\textwidth}  
  \vspace{-0.35cm}
  \centering
  \scalebox{0.95}{
  \includegraphics[
    width=1\linewidth
  ]{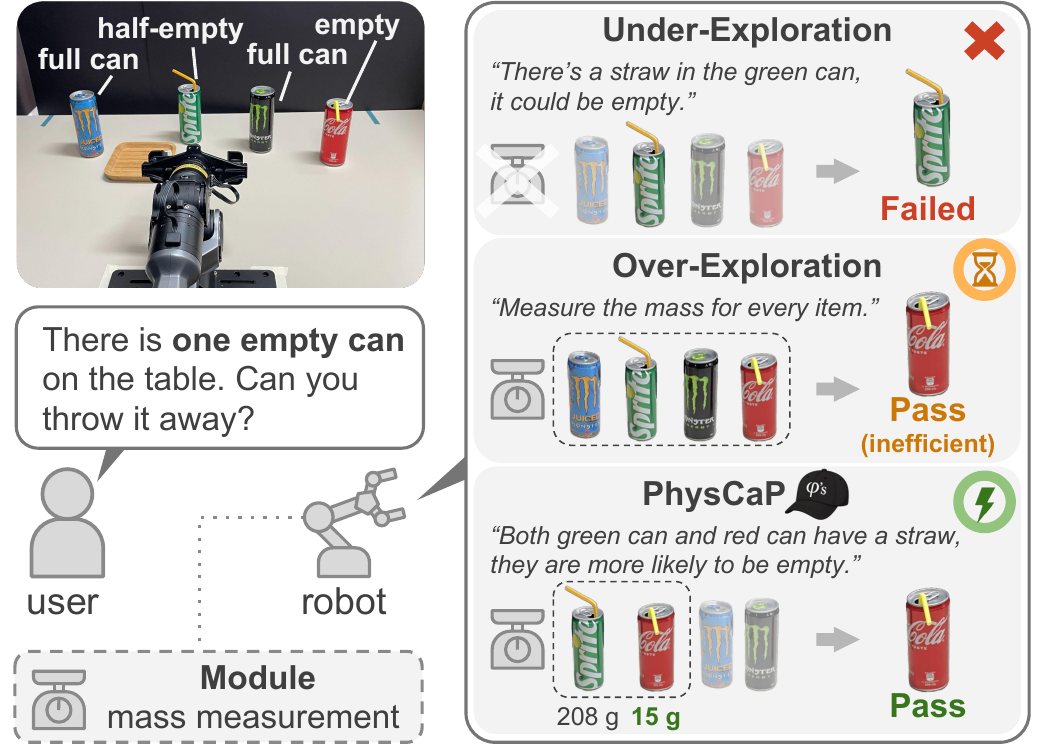}}
  \vspace{-0.1cm}
  \caption{
  \textbf{PhysCaP} augments Code-as-Policy (CaP) agents with physics-informed exploration, enabling inference of latent object properties (\eg mass) via physical property extraction modules to solve manipulation tasks requiring hidden-state estimation, \eg removing an empty can.
  }
  \label{fig:teaser}
  \vspace{-8pt}
\end{wrapfigure}

To extend code-as-policy agents \cite{liang2023oldcap}, \eg CaP-Agent0 \cite{fu2026cap}, with a physics-informed exploration layer for active perception, we propose the Physics-Informed Code-as-Policy (PhysCaP) agent, illustrated in \Cref{fig:teaser}. \method{} introduces physical property extraction modules that estimate latent physical properties such as mass and stiffness directly from robot proprioceptive feedback, without requiring additional training or sensing hardware \cite{wu2025savor} (\eg tactile sensors). To balance information sufficiency and exploration efficiency, we introduce a Planner to decide when exploration is necessary and when to terminate exploration, along with a Prioritizer that filters out implausible interactions and ranks the remaining candidates by a VLM-assigned priority score reflecting their likely task relevance. Overall, PhysCaP integrates code-as-policy reasoning with physics-informed exploration to enable efficient and generalizable grounding in the physical world.

We evaluate PhysCaP across three real-world tabletop manipulation tasks (identifying empty cans, selecting ripe avocados, and packing a grocery bag) using a 6-DoF AgileX PiPER robot arm, as well as a simulated empty-can task in LIBERO~\cite{liu2023libero} environment. While passive visual baselines (Code-as-Policy) and VLAs fail due to hidden physical properties and naive interactive methods over-explore, PhysCaP uses its Planner and Prioritizer to focus exploration on promising candidates, reducing both physical interactions and execution time. Furthermore, ablation studies demonstrate that VLMs cannot reliably infer hidden object properties \cite{wang2025phys2real, zhou2026physvlm}, motivating dedicated physical property extraction modules to accurately measure object mass and stiffness.

\vspacesection{Related Work}
\label{sec:related}

\textbf{Vision-language-action (VLA)} models leverage large-scale pretrained vision-language representations to directly map perceptual inputs and language instructions to low-level actions \cite{kim2024openvla, intelligence2025pi_, huang2025thinkact, fang2026molmoact2}. However, their reactive, feedforward nature often limits their ability to perform structured long-horizon reasoning, task decomposition, and explicit tracking of intermediate states \cite{huang2022innermonologue, shi2025hirobot}.

\textbf{Agentic robot learning frameworks} mitigate reasoning limitations by coordinating perception, planning, and control through code-as-policy execution \cite{huang2022innermonologue, song2023llmplanner, wang2023voyager, wang2024codeact, wang2025vlmseerobotdo}. These systems leverage foundation models \cite{team2024gemini} as callable modules, integrating vision-language and language-based reasoning into executable robot programs. However, they primarily focus on semantic and procedural reasoning \cite{zhao2023chat}, with limited capability to infer task-relevant physical properties. This motivates interaction-driven approaches that explicitly acquire missing physical information rather than relying solely on passive perception.

\textbf{Active perception and physical measurement} enable robots to reduce uncertainty through interaction in partially observable environments, evolving from reinforcement learning \cite{hu2025aawr} and VLM-based reasoning \cite{liu2024moka, jiang2024roboexp} to LLM-guided planning \cite{zhao2023chat, sun2024interactive, nazarczuk2025closed} and learned physical priors \cite{wang2025phys2real}. While recent methods achieve visuo-haptic \cite{wu2025savor} or proprioceptive estimation \cite{chen2025learning}, they typically decouple perception from task-level decision-making, lacking unified reasoning about what to measure and when to stop. In contrast, our proposed PhysCaP integrates proprioceptive estimation and Planner- and Prioritizer-guided exploration for efficient physical grounding, without requiring additional sensing hardware.

\vspacesection{PhysCaP: Physics-Informed Code-as-Policy Agent}
\label{sec:method}

\begin{figure}
    \centering
    \includegraphics[width=\linewidth]{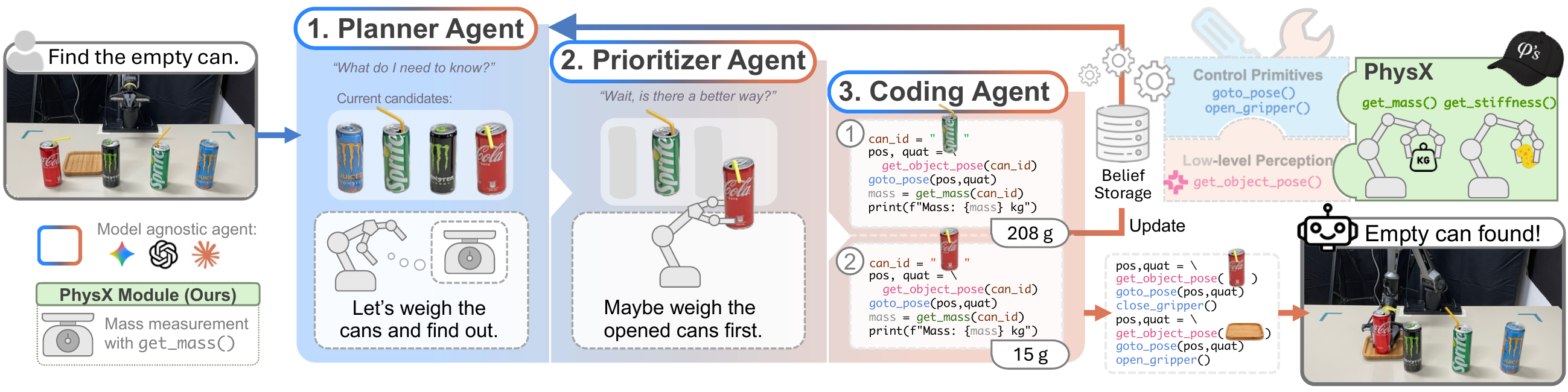}
    \caption{
    \textbf{Overview of PhysCaP, a Physics-Informed Code-as-Policy agent.} 
    Given a task requiring latent physical information, the Planner identifies what information is missing from the visual scene and proposes an initial exploration plan. The Prioritizer then refines this plan to improve interaction efficiency by filtering implausible actions and reducing redundant exploration. Finally, a code-generation agent produces executable programs that invoke the Physical Property Extraction (PhysX) modules (\texttt{get\_mass}, \texttt{get\_stiffness}) to actively measure the properties. The measurements update the system belief, which the Planner consults to decide whether to explore further or execute the task.
    }
    \label{fig:arch}
\end{figure}

We propose the Physics-Informed Code-as-Policy (PhysCaP) agent system (\Cref{fig:arch}), which targets partially observable manipulation settings where task success depends on latent physical properties that cannot be directly measured and no dedicated sensing hardware is available. PhysCaP actively interacts with the environment to acquire task-relevant physical information. Once sufficient evidence is collected, it uses the inferred properties to synthesize a more informed plan.

\subsection{Physical Property Extraction Modules and Control}

A natural approach is to use a VLM to infer latent physical properties from visual observations. However, as shown in our ablation studies, VLMs cannot reliably infer hidden physical properties from vision alone, and incorrect estimates can propagate errors to downstream reasoning and planning. We therefore adopt an interaction-driven approach, where the robot actively measures otherwise unobservable properties. We introduce two training-free physical property extraction (PhysX) modules for estimating object mass (\Cref{sec:mass-measurement}) and stiffness (\Cref{sec:stiffness-measurement}) using robot proprioception and standard gripper hardware (see \Cref{fig:task}). The low-level control API is described in \Cref{sec:low-level}.

\subsubsection{Object Mass Measurement}
\label{sec:mass-measurement}

\begin{figure}[t]
    \centering
    \includegraphics[width=\linewidth]{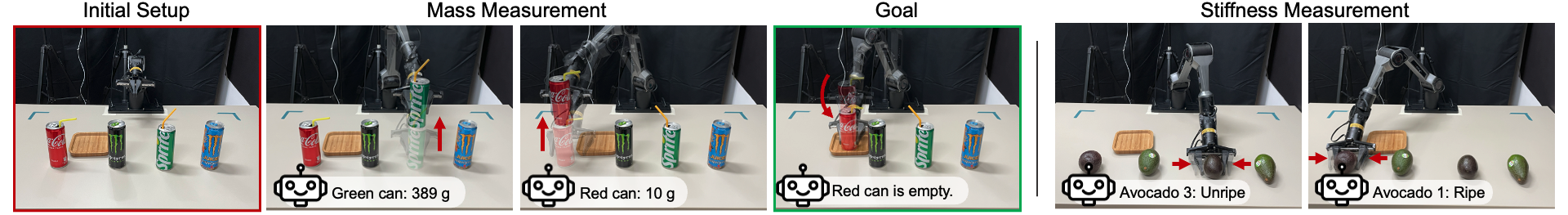}
    \caption{\textbf{PhysCaP measures task-relevant latent physical properties beyond passive visual perception and completes tasks.}
    PhysCaP decides which tools (\texttt{get\_mass}, \texttt{get\_stiffness}) to invoke, then synthesizes an informed code policy grounded in those measurements (left: Identify Empty Can; right: Pick Ripe Avocado).}
    \label{fig:task}
\end{figure}

Our mass measurement module (\texttt{get\_mass}) estimates object mass from joint torques. The robot performs a fixed 15 cm lift, recording torques in an empty grasp and while holding the object, with brief pauses to reduce vibration (\Cref{fig:task}). The differential torque $\Delta\tau=\tau_{loaded}-\tau_{empty}$ isolates the object's gravity contribution, and
$\hat{m}=\mathrm{J}_z\cdot\Delta\tau/(g|\mathrm{J}_z|^2)$,
where $\mathrm{J}_z$ is the vertically projected end-effector Jacobian and $g=9.8~\mathrm{m/s^2}$. This formulation is pose-robust to placement variations.

\subsubsection{Object Stiffness Measurement}
\label{sec:stiffness-measurement}

Our stiffness measurement module (\texttt{get\_stiffness}) estimates object rigidity from gripper displacement $d$ and force $F$. The gripper closes incrementally until contact, then performs a brief backoff test, in which a force decrease $\Delta F$ upon reopening confirms contact and records the reference displacement $d_0$. It then continues closing to a target force $F^*$ while recording $(d_i,F_i)$. Stiffness is estimated by ordinary least squares as $k=\Delta F/\Delta d~\mathrm{(N/m)}$, then mapped to five levels $s\in\{1,\dots,5\}$ from Ultra-Soft to Rigid. Each object is measured five times, with the final level selected by majority vote.

\subsubsection{Low-level Robot Control}
\label{sec:low-level}

We build our control framework on CaP-Agent0~\cite{fu2026cap} to ground high-level agentic reasoning in real-world manipulation. The system provides modular APIs, including \texttt{get\_object\_pose}, \texttt{goto\_pose}, \texttt{open\_gripper}, and \texttt{close\_gripper}, using Molmo2~\cite{clark2026molmo2} for object localization and ZED 2i for depth estimation. These APIs abstract high-level reasoning from embodiment-specific control, enabling deployment across robot platforms and supporting closed-loop physical exploration. Details are provided in \Cref{sec:app_physx,sec:app_control}.

\subsection{Planner Agent}
\label{sec:planner}

Although VLMs reason well over observable scenes, they remain blind to latent physical properties, causing passive agents to act with insufficient context. We introduce a Planner that assesses whether current observations suffice for reliable task execution, identifies missing physical information, and triggers targeted exploration when needed. It stops exploration once sufficient evidence is gathered and invokes the downstream code-generation agent, enabling efficient interaction-driven reasoning while avoiding \textit{over-exploration}. Prompt details and exploration example are provided in \Cref{sec:app_planner}.

\subsection{Prioritizer Agent}
\label{sec:prioritizer}

While the Planner decides when to explore, naive interactions may cause \textit{over-exploration} and unnecessary cost. We introduce a Prioritizer that refines the Planner’s candidate interactions using visual heuristics, filtering implausible or redundant plans and ranking the rest by priority. For example, it may prioritize inspecting a potentially ripe avocado (black) over clearly unripe ones (green). The Prioritizer assigns each candidate a priority score, briefly justifies the score, and reorders candidates accordingly, enabling task-relevant information acquisition with minimal interaction cost. Detailed prompts and examples are provided in \Cref{sec:app_prioritizer}.

\vspacesection{Experiments}
\label{sec:experiment}

\vspacesubsection{Experimental Setup}
\label{sec:setup}

\textbf{Hardware Setup.} 
Our setup consists of a 6-DoF AgileX PiPER robotic arm mounted on a height-adjustable table. 
A ZED 2i depth camera is positioned above and slightly behind the arm to observe the workspace. Detailed hardware specifications and configurations are provided in \Cref{sec:app_exp_details}.

\textbf{Foundation Models.} All agents (detailed in \Cref{sec:planner,sec:prioritizer}) use Gemini 3.1 Pro as the primary backbone, while object grounding for \texttt{get\_object\_pose} relies on Molmo2 \cite{clark2026molmo2}. To ensure a fair comparison, all baseline methods share the identical backbone. However, our framework is inherently model-agnostic.

\textbf{Evaluation Protocol and Metrics.}
We quantitatively evaluate our framework and baselines over 10 trials per task using three metrics. Task success rate (SR $\uparrow$) measures the percentage of tasks completed successfully, while Objects Interacted (OI $\downarrow$) counts physical exploration actions before task completion, including repeated interactions with the same object, and Execution Time (Time $\downarrow$) measures total robot execution time. Higher efficiency is reflected by fewer interactions and shorter execution times. OI and Time are reported only for successful episodes. More details (\eg object pose perturbations) can be found in \Cref{sec:app_exp_details}.

\vspacesubsection{Tasks}
\label{sec:tasks}

\begin{figure}
    \centering
    \includegraphics[width=\linewidth]{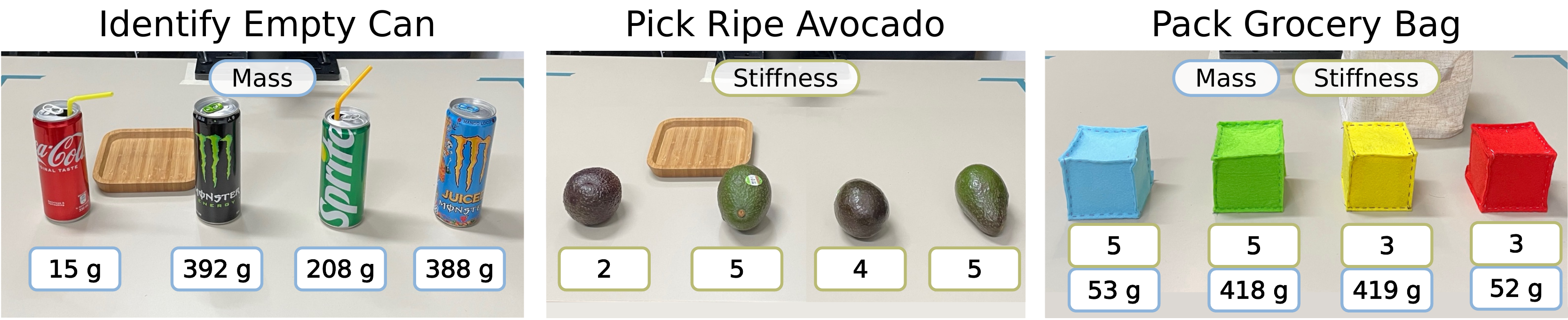}
    \caption{
    \textbf{Ground-truth physical properties} of the relevant target objects are annotated in each task setup, with mass (g) and/or stiffness ($s \in \{1,\dots,5\}$) shown according to properties required by each tasks.}
    \label{fig:task_visualize}
\end{figure}

We evaluate PhysCaP on three partially observable tabletop manipulation tasks requiring efficient physical exploration (\Cref{fig:task_visualize}). Using standardized objects and 3D-printed cubes, we vary visual, semantic, and behavioral factors following \cite{gao2026taxonomy}, including object positions/colors, target referring expressions, and hidden physical properties. Implementation details, prompts, hardware, and task specifications are provided in \Cref{sec:app_task_details}.

\textbf{Task 1: Identify Empty Can.} 
The goal is to identify the single empty soda can among four cans and place it on a wooden tray. The scene contains two sealed cans and two open cans with inserted straws. This task evaluates whether the agent can combine semantic priors with targeted physical interaction. A sealed can is unlikely to be empty, so an effective agent should treat the two opened cans as the promising candidates and use \texttt{get\_mass} to confirm which one is empty, rather than weighing every can on the table.

\textbf{Task 2: Pick Ripe Avocado.} 
The goal is to identify the single ripe avocado among four and place it on a wooden tray. The scene contains two green avocados that are unripe and two dark avocados that both appear ripe, though only one is. This task evaluates whether the agent can use visual cues to narrow the search space and then settle the remaining ambiguity through physical measurement. Since skin color cannot separate the two dark candidates, an effective agent should skip the green avocados and use \texttt{get\_stiffness} to determine which of the two is actually ripe.

\textbf{Task 3: Pack Grocery Bag.}
The goal is to select a suitable item as the base of a grocery bag. The heavier and stiffer objects generally provide greater stacking stability.
Since the cubes look identical in shape, an agent cannot infer hidden mass and stiffness from vision alone; however, it would also not be ideal to test both properties for all cubes. An effective agent should eliminate two lighter candidates and only measure the stiffness of the heavier cubes after their mass has been measured.

\vspacesubsection{Baselines and Variants}
\label{sec:baselines}

We systematically evaluate our full system against three progressively capable baselines to isolate the contribution of each module. \textbf{CaP} (CaP-Agent0 in \cite{fu2026cap}) relies solely on visual reasoning and has no physical exploration capability, exposing the failure modes of purely passive perception under hidden physical properties. \textbf{CaP+PhysX} augments this baseline with physical property extraction (PhysX) modules, enabling measurement of latent properties but requiring exhaustive interaction with all objects due to the absence of exploration reasoning. \textbf{CaP+PhysX+Planner} further adds a VLM-based planner that introduces a stopping criterion for exploration, but interactions are still selected randomly. \textbf{PhysCaP}, our full model, additionally incorporates the Prioritizer, which eliminates implausible interactions and orders the remaining ones by a visual-heuristic priority score, enabling efficient and targeted physical reasoning.

\vspacesubsection{Real-World Task Performance}
\label{sec:results}

\begin{table*}[t]
    \centering
    \caption{\textbf{Task performance.}
    PhysCaP achieves the lowest/tied interaction count across all tasks and the shortest execution time on two of three tasks. OI ($\downarrow$) denotes the average number of physical object interactions during exploration including repeated interactions with the same object, while Time ($\downarrow$) reports the total robot execution time. }  
    \vspace{0.1cm}
    \resizebox{\textwidth}{!}{%
    \begin{tabular}{l cc cc cc}
        \toprule
        \multirow{2}{*}{\textbf{Method}} & 
        \multicolumn{2}{c}{\textbf{Task 1: Identify Empty Can}} & 
        \multicolumn{2}{c}{\textbf{Task 2: Pick Ripe Avocado}} & 
        \multicolumn{2}{c}{\textbf{Task 3: Pack Grocery Bag}} \\
        \cmidrule(lr){2-3} \cmidrule(lr){4-5} \cmidrule(lr){6-7} 
        & 
        OI ($\downarrow$) & Time ($\downarrow$) & 
        OI ($\downarrow$) & Time ($\downarrow$) & 
        OI ($\downarrow$) & Time ($\downarrow$) \\
        \midrule

        CaP+PhysX & 
        $4.00$ & $396.84 \pm 15$ s & 
        $4.00$ & $515.61 \pm 9$ s &
        $9.17$ & $1019.32\pm 376$ s \\ 
        
        CaP+PhysX+Planner &  
        $3.75$ & $368.40 \pm 62$ s & 
        $4.13$ & $563.29 \pm 76$ s &
        $7.80$ & $853.14\pm 54$ s \\ 

        PhysCaP-joint  & 
        $\mathbf{2.00}$ & $\mathbf{222.11 \pm 2}$ s & 
        $2.71$ & $384.39 \pm 111$ s &
        $6.50$ & $682.3\pm 139$ s \\ 
        
        PhysCaP (Ours) & 
        $\mathbf{2.00}$ & $256.57 \pm 62 \text{ s}$ & 
        $\mathbf{2.00}$ & $\mathbf{300.47 \pm 53 \text{ s}}$ &
        $\mathbf{5.50}$ & $\mathbf{572.79\pm 61}$ s \\ 
        
        \bottomrule
    \end{tabular}%
    }
    \label{tab:main-results}
\end{table*}

\begin{figure}[t]
    \centering
    \begin{subfigure}[b]{0.47\linewidth}
        \centering
        \includegraphics[width=\linewidth]{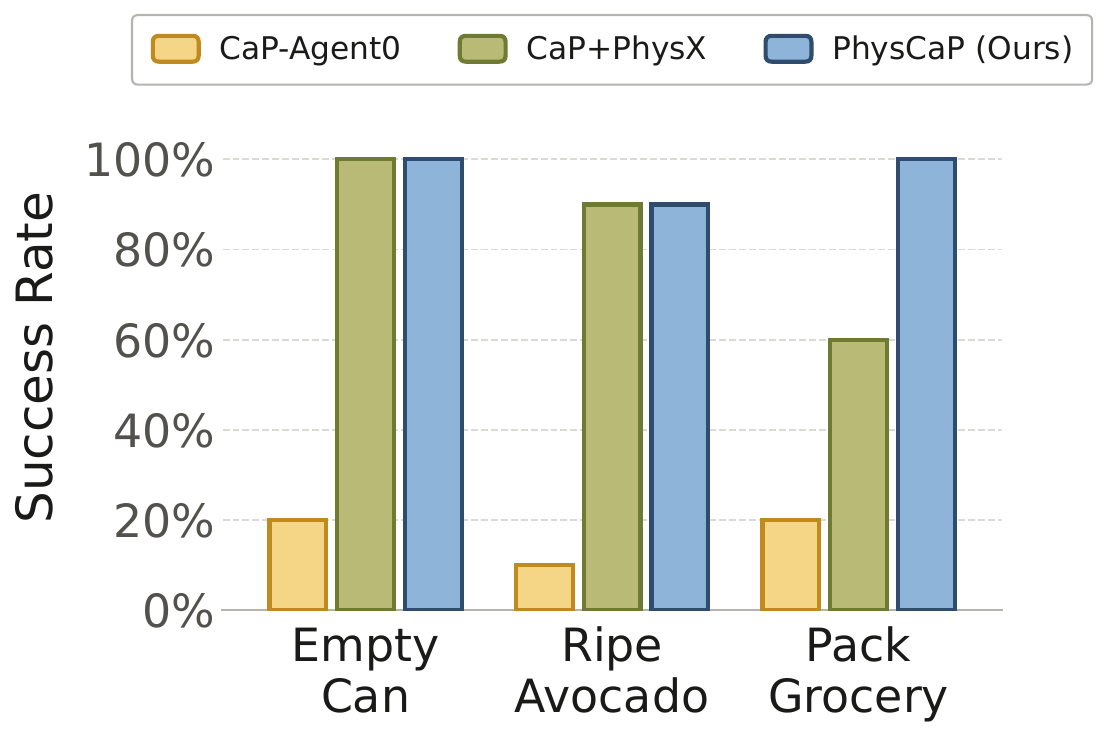}
        \caption{Success Rate ($\uparrow$)}
        \label{fig:results-sr}
    \end{subfigure}
    \hfill
    \begin{subfigure}[b]{0.47\linewidth}
        \centering
        \includegraphics[width=\linewidth]{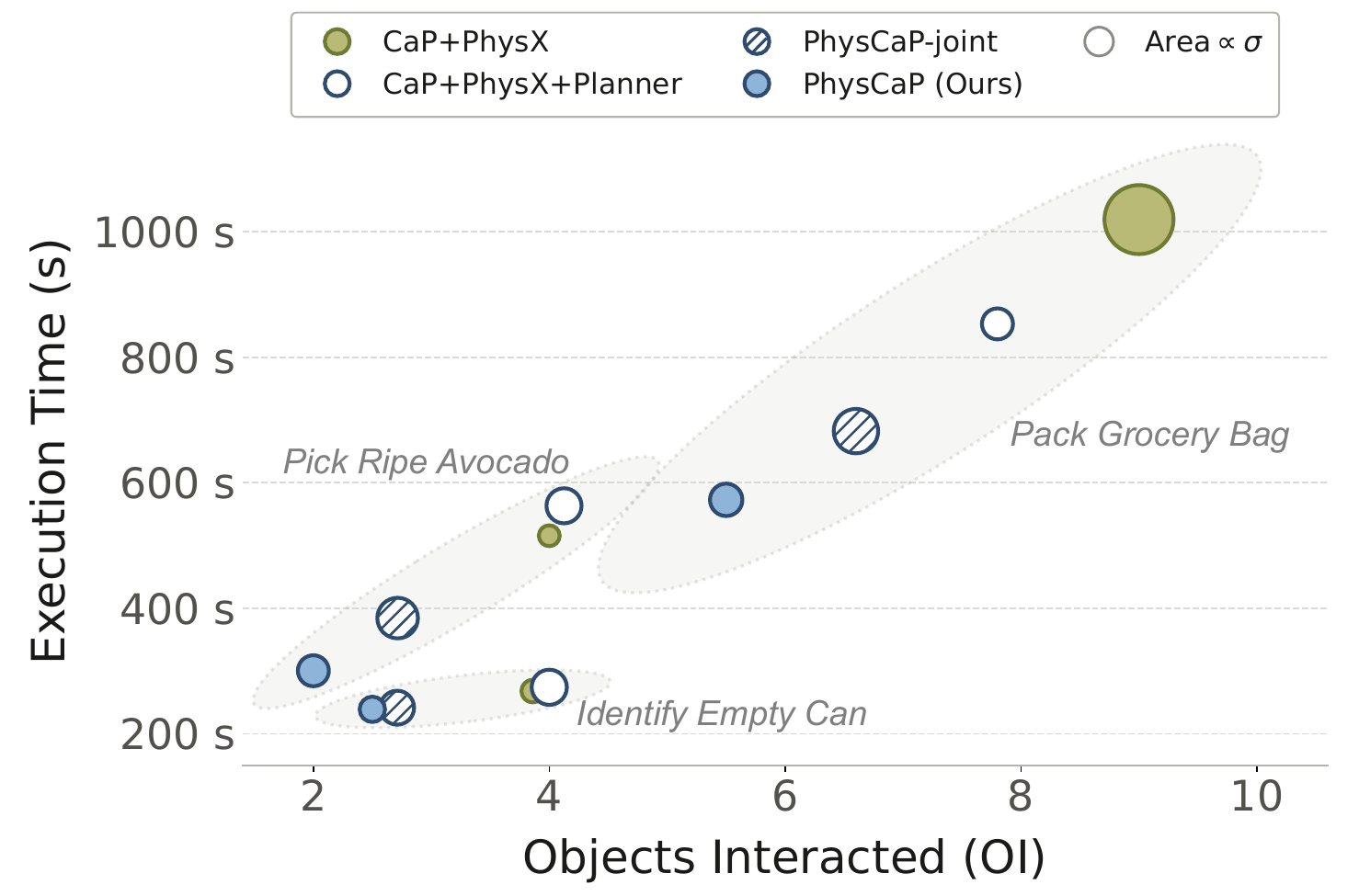}
        \caption{Exploratory Efficiency ($\downarrow$)}
        \label{fig:results-eff}
    \end{subfigure}
    \caption{\textbf{Real-world task performance and exploratory efficiency.}
    (a) PhysCaP matches CaP+PhysX on Empty Can and Ripe Avocado, and improves on the longer-horizon Pack Grocery, where CaP+PhysX can lose track of prior measurements, leading to remeasurement or failure.
    (b) Efficiency in physical exploratory interactions and total robot execution time; bubbles are grouped by task, with area indicating standard deviation. Lower-left is more efficient.}
    \label{fig:results-main}
\end{figure}

\Cref{fig:results-sr} shows that vision-only \textbf{CaP} performs poorly because it lacks hidden physical properties. \textbf{CaP+PhysX} provides these estimates but explores objects in scene order, resulting in low efficiency (see \Cref{fig:results-eff}). Adding a VLM-based Planner (\textbf{CaP+PhysX+Planner}) enables closed-loop stopping, but \Cref{tab:main-results} shows that stopping alone does not consistently improve efficiency because interactions remain unprioritized, and localization ambiguity can also trigger redundant measurements. We therefore test joint reasoning in \textbf{PhysCaP-joint}, where one VLM call jointly determines when to stop and which object to inspect next. Although more efficient than the Planner alone, its behavior remains inconsistent across trials, sometimes relying on visual cues and otherwise exploring nearly exhaustively. This motivates \textbf{PhysCaP}, which separates the roles by having the Planner determine when to stop while a Prioritizer filters and ranks candidates. This separation yields more consistent selective exploration, with the lowest or tied OI across tasks and the lowest Time on two tasks (\Cref{tab:main-results}). The generated code examples are detailed in \Cref{sec:app_code_demo}.

\vspacesubsection{Comparisons to VLAs in Simulation}

\begin{wrapfigure}{r}{0.5\textwidth}
    \vspace{-0.4cm}
    \centering
    \includegraphics[width=\linewidth]{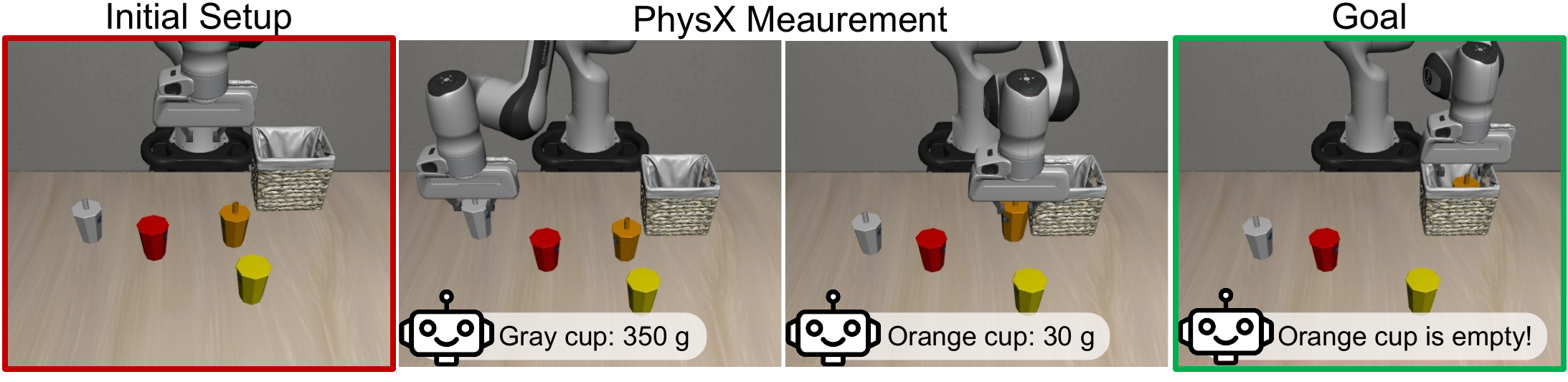}
    \vspace{-0.5cm}
    \captionof{figure}{\textbf{Replicate Identify Empty Can task in the LIBERO environment.} 
    }
    \label{fig:results-sim}

    \vspace{0.3cm}

    \captionof{table}{\textbf{Results of simulated Identify Empty Can task}.
    }
    \label{tab:sim_results}
    \scalebox{0.82}{%
    \begin{tabular}{l c c c}
    \toprule
    \textbf{Method} & 
    SR ($\uparrow$) & 
    OI ($\downarrow$) & 
    Time ($\downarrow$) \\
    
    \midrule

    CaP+PhysX & 
    $74\%$ & $2.16$ & $\mathbf{70.81} \pm 32.50$  s \\
    
    CaP+PhysX+Planner & 
    $62\%$ & $1.45$ & $79.55 \pm 44.49$ s \\

    PhysCaP (Ours) & 
    $\mathbf{78\%}$ & $\mathbf{1.44}$ & $71.24 \pm 18.63$ s \\
    
    \midrule
    
    OpenVLA & 
    $0\%$ & \textemdash & \textemdash \\
    
    $\pi_{0.5}$ & 
    $4\%$ & 1.50 & \textemdash \\
    
    MolmoAct2 & 
    $23\%$ & $1.04$ & \textemdash \\
    
    \bottomrule
    \end{tabular}
    }
    \vspace{-0.4cm}
\end{wrapfigure}

All the baselines so far are code-as-policy agents, like PhysCaP. We now compare against vision-language-action policies, which map observations to actions without an explicit exploration step.
Since it is non-trivial to directly adapt VLA DROID checkpoints to our PiPER setup, we run this comparison in simulation using the LIBERO environment~\cite{liu2023libero, zhou2025liberopro}.

\textbf{Task.} 
We replicate the Identify Empty Can task in LIBERO. 
As illustrated in \Cref{fig:results-sim}, the scene contains four cups (one empty and three full) and a target basket, with randomized cup poses and perturbed object mass. As the empty cup is not visually identifiable, the agent must rely on physical interaction to infer its state.

\textbf{Evaluation Metrics.} 
We use the same evaluation metrics as in the real-world setting (SR $\uparrow$, OI $\downarrow$, and Time $\downarrow$), with execution time computed as $N_{\text{sim\_steps}} / 20\,\text{Hz}$.

\textbf{Baselines.} We compare against various VLA models, including OpenVLA~\cite{kim2024openvla}, $\pi_{0.5}$\cite{intelligence2025pi_}, and MolmoAct2~\cite{fang2026molmoact2}, using their publicly released LIBERO checkpoints.

\textbf{Physical Property Extraction Module in Simulation.} 
To align with our real-world PhysX \texttt{get\_mass} module, we implement a physical sensing API in simulation where mass is not directly observable; instead, ground-truth mass is revealed only after a verified interaction: the agent must grasp the cup for at least 0.5 seconds and lift it 3 cm above its reference height for 0.5 seconds. This design enforces explicit physical exploration to acquire information.

\textbf{Results and Analysis.} 
The results (\Cref{tab:sim_results}) closely match real-world trends. PhysCaP achieves higher success with fewer object interactions than CaP+PhysX, while execution times remain comparable because additional physical measurements add little execution time in simulation. In contrast, most VLAs fail, achieving near-zero success rates. These models directly map vision to actions without mechanisms for hypothesis-driven exploration, resulting in failure when latent physical uncertainty is present. Note that OI and Time are reported only for successful episodes across 50 trials, and are therefore undefined for OpenVLA, while $\pi_{0.5}$ and MolmoAct2 typically select a can randomly, resulting in low SR and correspondingly low OI.

\vspacesection{Discussion}

\begin{figure}[t]
    \centering
    \begin{subfigure}[b]{0.46\linewidth}
        \centering
        \includegraphics[width=\linewidth]{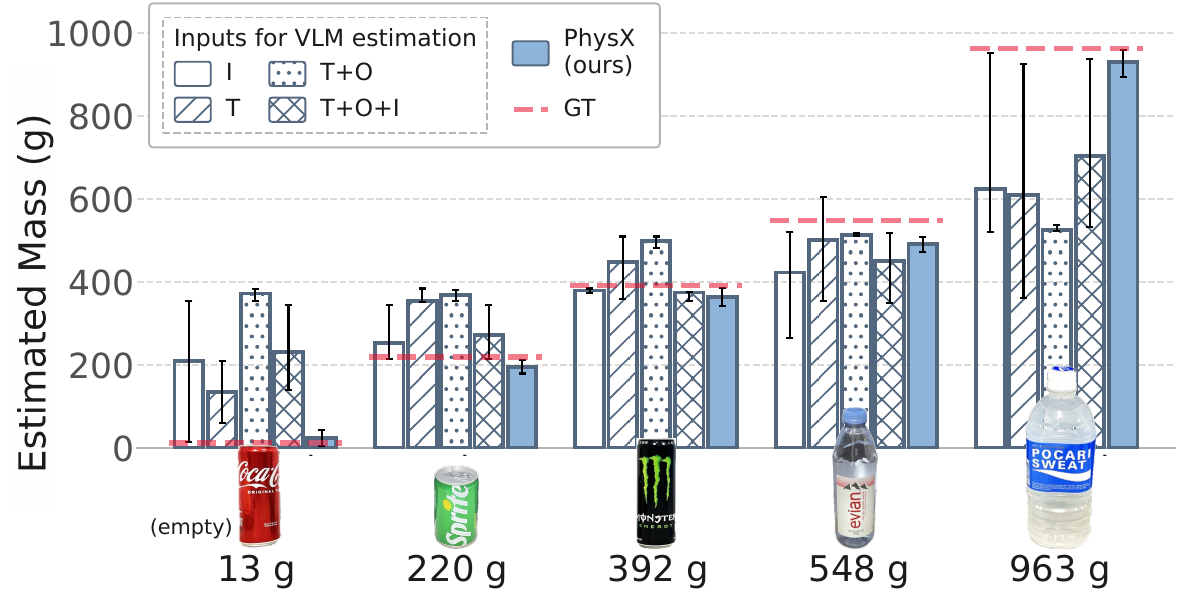}
        \caption{Mass estimation ($n=256$)}
        \label{fig:phys-mass}
    \end{subfigure}
    \hfill
    \begin{subfigure}[b]{0.52\linewidth}
        \centering
        \includegraphics[width=\linewidth]{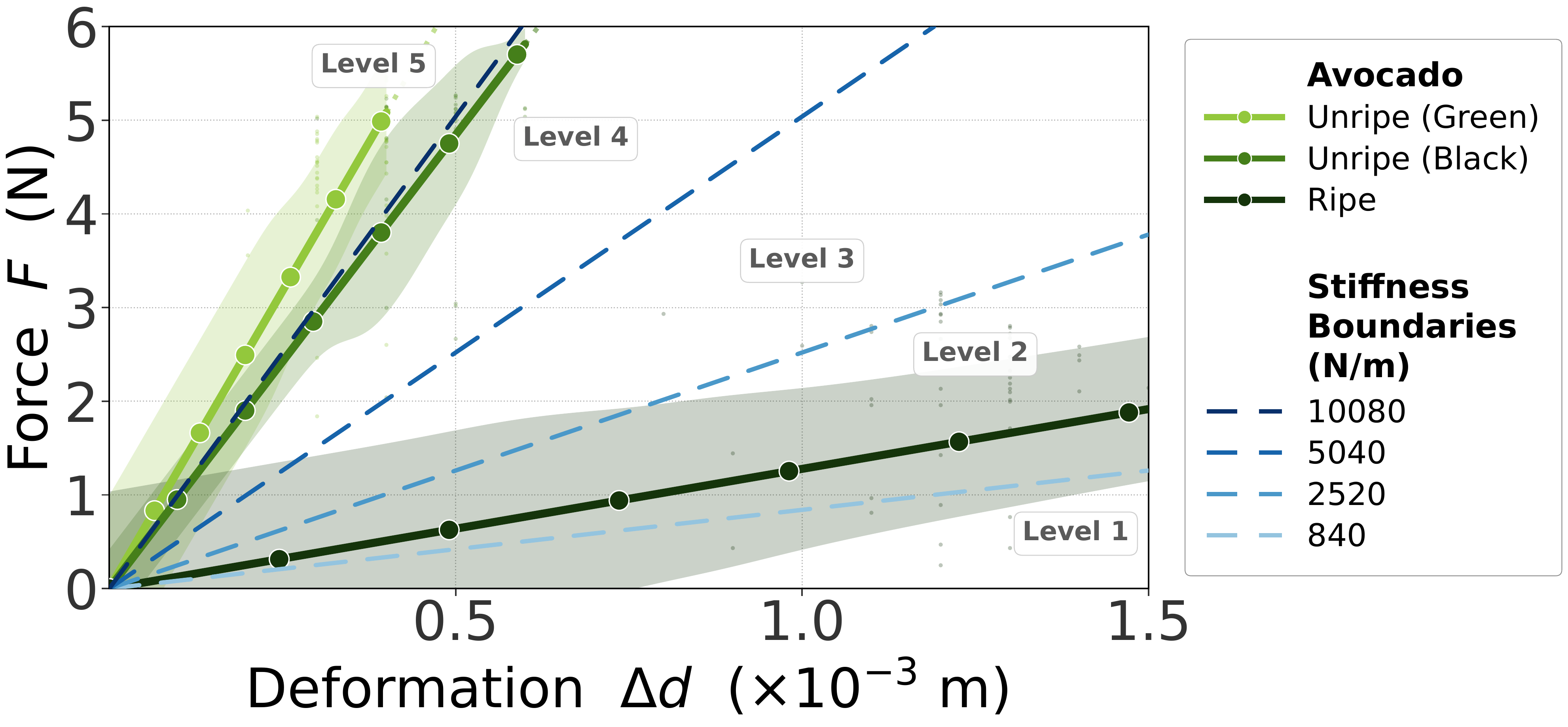}
        \caption{Stiffness from force--displacement}
        \label{fig:phys-stiffness}
    \end{subfigure}
    \caption{\textbf{Validating the PhysX modules against known ground truth.}
    (a) Across five objects spanning 13--963\,g, PhysX tracks ground truth (GT) over the whole range, while VLM estimates---given the reference image (I), raw torque (T), and object name (O) in combination---are misled by appearance, \eg substantially overestimating the empty can.
    (b) Squeezing three avocados of differing ripeness: green lines are linear fits with shading for measurement variance, and dashed blue lines mark the stiffness-level boundaries set by the calibrated spring reference. The ripe avocado (Level 2) separates cleanly from both unripe ones (Levels 4--5).}
    \label{fig:phys-eval-merged}
\end{figure}

\vspacesubsection{Limitations of VLMs in Physical Property Estimation}

VLMs inherit a great deal of human knowledge about everyday objects and can produce roughly plausible mass estimates, yet \Cref{sec:results} shows that they struggle with tasks requiring hidden physical properties. We therefore ask what information enables VLMs to solve such tasks.

We progressively augment the VLM input with the reference image (\textbf{I}), raw joint torque (\textbf{T}), and object name (\textbf{O}). The image captures naturally available visual information, while torque provides physical evidence analogous to the readings used by our \texttt{get\_mass} PhysX module. Adding the object name further removes ambiguity about the lifted object. On the five reference objects in \Cref{fig:phys-mass}, we report mean absolute error (MAE), the average absolute difference between predicted and ground-truth mass. The image alone yields an MAE of $151.5 \pm 182.5$ g ($n=755$). Adding torque ($193.0 \pm 189.7$ g), object identity ($217.4 \pm 155.8$ g), or both ($134.5 \pm 148.4$ g) does not significantly improve accuracy over the image alone ($t=0.93$).

These results suggest that the limitation is not simply missing information, but the VLM's difficulty in converting physical measurements into accurate quantitative estimates. This motivates our PhysX modules, which explicitly extract physical properties from robot sensor measurements rather than relying on VLM reasoning alone.

\vspacesubsection{Validating the Physical Property Extraction Modules}

We now evaluate how precisely the PhysX modules measure physical properties. For each module we construct a reference with ground-truth values, so we can quantify the measurement error.

\textbf{Mass Measurement.} Across the same five reference objects spanning 13 g to 963 g ($n=250$ in \Cref{fig:phys-mass}), PhysX achieves $38.1 \pm 26.0$ g MAE, outperforming the best VLM condition ($t=5.6$). Its lower standard deviation also indicates more consistent measurements, with errors remaining stable across object loads. In Identify Empty Can, the empty and lightest full can differ by $193$ g. PhysX's error remains well within this margin, whereas VLM errors of $134.5$ to $217.4$ g make reliable identification difficult.

\textbf{Stiffness Measurement.} We evaluate the stiffness module on a calibrated 3D-printed button and real avocados. The button uses interchangeable springs with known stiffness ($k=420$ N/m each), defining four boundaries and five reference levels from Ultra-Soft (1) to Rigid (5) that span our task objects. We squeeze three avocados of different ripeness 50 times each and record gripper deformation and force. \Cref{fig:phys-stiffness} shows the ripe avocado at Level 2, while the unripe ones fall into Levels 4 and 5. Their measurements do not overlap with the ripe avocado, so a single squeeze distinguishes them, whereas the two unripe avocados overlap as expected.

\vspacesubsection{Robustness Across VLM Backbones}
\begin{wrapfigure}{r}{0.55\textwidth}
    \vspace{-0.4cm}
    \centering
    \includegraphics[width=0.47\linewidth]{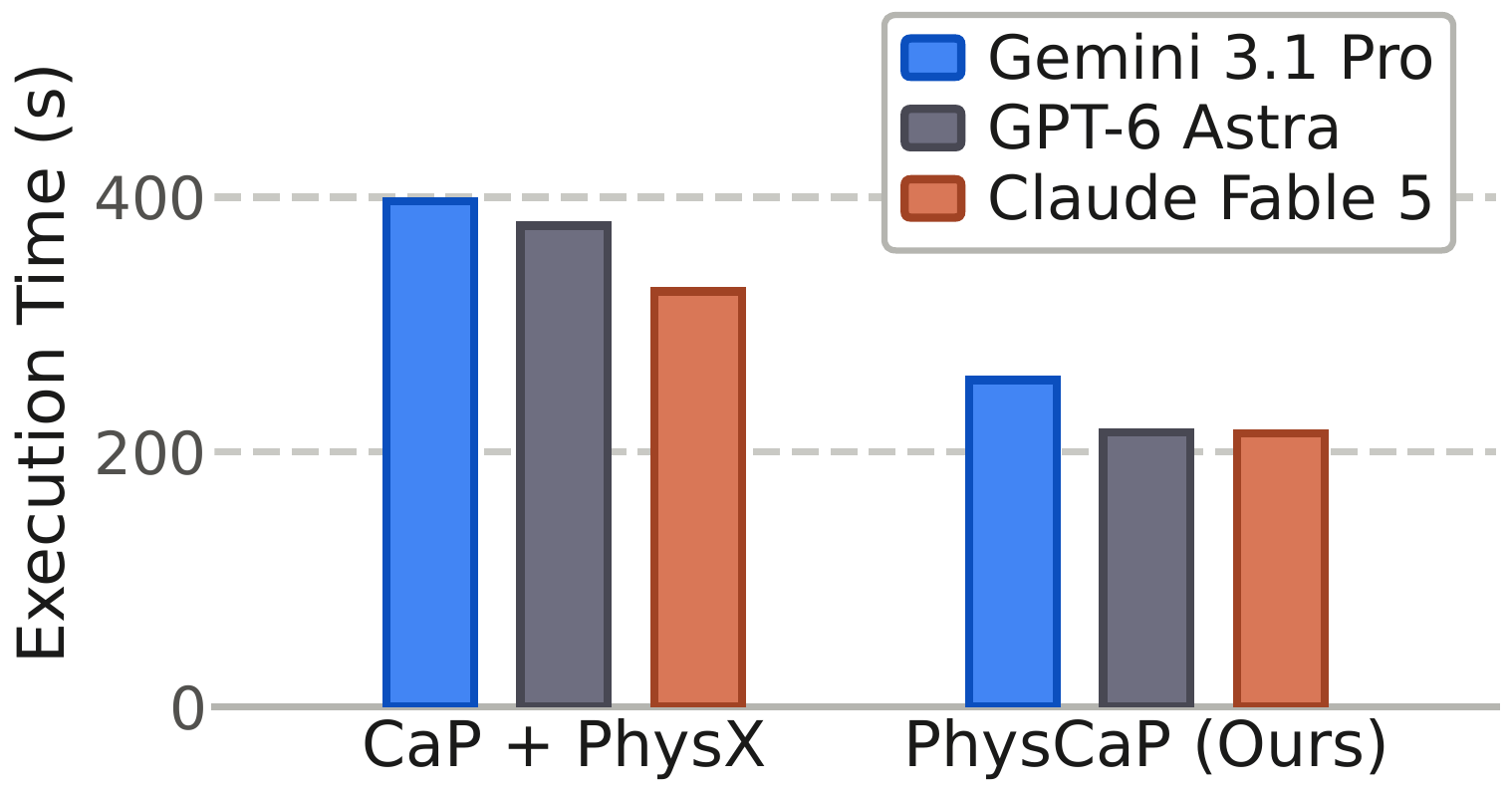}
    \hfill
    \includegraphics[width=0.47\linewidth]{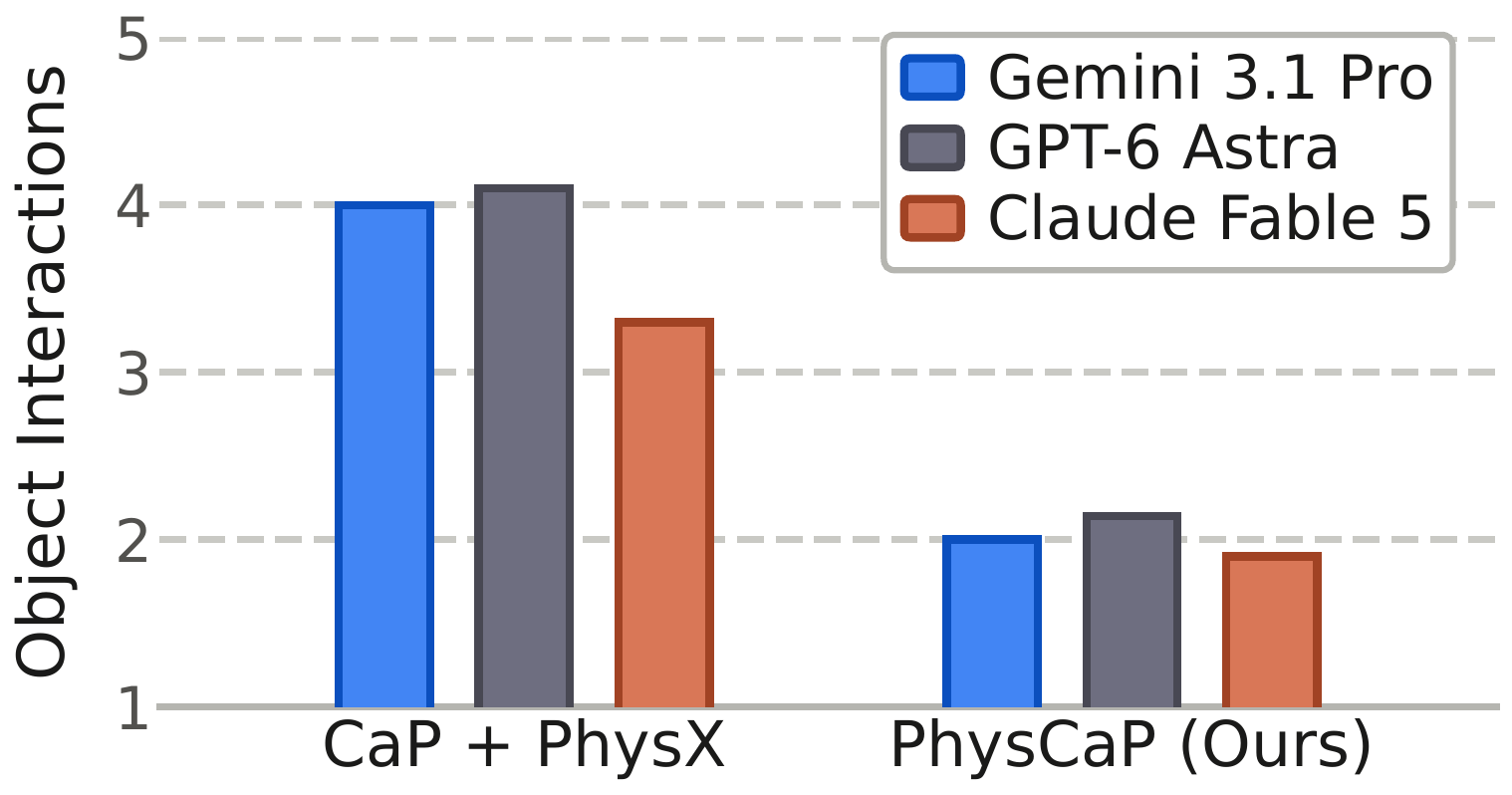}
    \caption{\textbf{Cross-VLM generalization.}
    We test three VLM backbones on Identify Empty Can. Across all three, PhysCaP consistently outperforms the CaP+PhysX baseline on execution time (left) and object interactions (right), showing that our framework generalizes across VLM backbones.}
    \label{fig:models}
\end{wrapfigure}

PhysCaP is an agentic system, so its backbone can in principle be swapped for any sufficiently capable model. We repeat Identify Empty Can with Gemini 3.1 Pro, Claude Fable 5, and GPT-6 Astra as backbones for the Planner, Prioritizer, and coding agent. All configurations achieve 100\% success, so we compare exploratory efficiency using OI and execution time. PhysCaP consistently uses fewer interactions and less time than CaP+PhysX across backbones. Notably, Gemini 3.1 Pro achieves OI $2.0$, compared with $4.1$ and $3.3$ for CaP+PhysX with GPT-6 Astra and Claude Fable 5, respectively, indicating that the efficiency gain comes from the architecture rather than the backbone.

\vspacesection{Conclusion}
\label{sec:conclusion}

We introduced PhysCaP, a physics-informed code-as-policy agent system that actively measures the physical properties a task depends on instead of guessing them from appearance. PhysX estimates mass and stiffness from proprioception alone, while a Planner and Prioritizer decide when to explore and what to interact with. Across three real-world tasks and a simulated LIBERO task, PhysCaP  matches or exceeds on every task requiring physical interaction, with no meaningful loss on the visual-only task. Our ablations show why both parts matter. A VLM's mass estimates are too coarse to tell an empty can from a full one, no matter what inputs it is given, while an explicit physics formula reduces the error by more than half. Prioritization helps when merged into the Planner's prompt, but a single call applies its own heuristics inconsistently, whereas a dedicated agent recovers the remaining efficiency. Explicit physical grounding and explicit agent role separation are therefore complementary, with one supplying information the agent cannot see, the other keeping the cost of acquiring it low.

\textbf{Limitations.} While PhysCaP enables generalizable physical reasoning \cite{fu2026cap}, the current implementation remains limited by model variability, perception accuracy, and hardware execution. Specifically, commercial VLM APIs \cite{team2024gemini}, 2D localization \cite{clark2026molmo2} mapped to single-camera depth, and communication latency with the PiPER arm can respectively affect reasoning consistency, 3D end-effector precision, and trajectory execution. \Cref{sec:failure_modes} discusses these limitations in detail, together with their associated failure modes and potential improvements.

\bibliography{main}  

\clearpage
\appendix
\onecolumn

\vspace{-1.5cm}

\doparttoc
\faketableofcontents

\begingroup
\hypersetup{linkcolor=black}
\hypersetup{pdfborder={0 0 0}}
\part{Appendix} 
\parttoc 
\endgroup

\textbf{Project website.} We present qualitative results (robot execution and generated codes) on our project website: \href{https://physcap.github.io/}{https://physcap.github.io/}.

\section{Experiment Details}
\label{sec:app_exp_details}

\begin{figure}[t]
  \centering
  \scalebox{1}{
  \includegraphics[
    width=1\linewidth
  ]{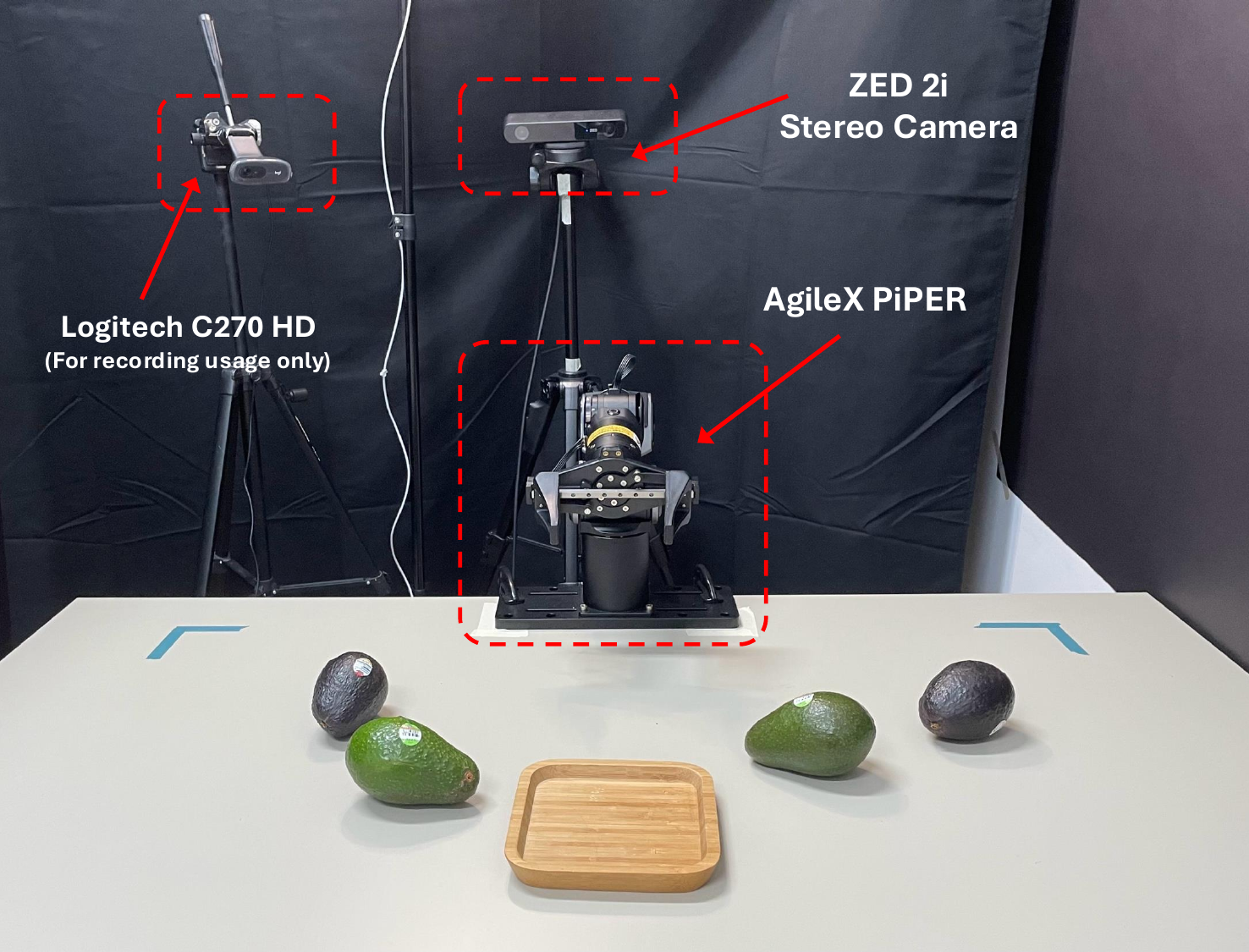}}
  \caption{\textbf{Our experiment setup.} The ZED 2i stereo camera is mounted directly behind the AgileX PiPER robot arm to provide a perspective that aligns with the robot's view of the manipulation area. A supplementary Logitech C270 HD is placed to the side to capture external footage for recording purposes.
  }
  \label{fig:setup}
\end{figure}

\textbf{Experiment Setup.} As depicted in \Cref{fig:setup}, the physical platform comprises an AgileX PiPER 6-DoF robotic arm equipped with a gripper, mounted on a height-adjustable tabletop ($120 \text{cm} \times 70 \text{cm}$) workspace. Spatial perception is provided by a side-mounted ZED 2i stereo camera capturing RGB-D images at a resolution of $1280 \times 720$. All computation, including visual processing and policy inference, is executed on a workstation equipped with an RTX 3090 GPU (24 GB VRAM).

\textbf{Visual Perception Pipeline.}
We utilize the side-mounted ZED 2i stereo camera to capture global RGB scene observations and corresponding depth maps. Object semantic localization is handled by the high-level reasoning agent, which passes the target visual context to a pre-trained Molmo2 model \cite{clark2026molmo2} for 2D image coordinate pointing. During inference, once Molmo2 outputs the exact 2D pixel coordinates, the system back-projects the point using the ZED depth map and applies the camera's extrinsic transformation matrix to resolve the absolute 3D world coordinate. This spatial location is then parameterized and passed to the \texttt{get\_object\_pose()} control API to guide downstream joint trajectories. A detailed breakdown of the low-level mechanical routines governed by \texttt{get\_object\_pose()} is provided in \Cref{sec:app_control}.

\subsection{Real-World Task Details}
\label{sec:app_task_details}

This section provides comprehensive implementation details, task setup configurations, and evaluation criteria for the three real-world tabletop manipulation tasks described in \Cref{sec:tasks}. For each task, the interactable objects will be placed in a similar layout, while each object's initial position will be randomized with slight variation across trials.

\subsubsection{Task 1: Identify Empty Can}
\label{sec:app_task_can}

\textbf{Scene Description.} The workspace contains four soda cans of different colors, as shown in \Cref{fig:task}. Two of the cans (blue and black) are unopened and full of liquid, while the other two (red and green) are visibly open and contain straws. Of the two open cans, one (red) is completely empty, whereas the other (green) still contains liquid (weighing over $100 \text{g}$). A wooden tray is provided for the robot to place the target can.

\textbf{Objective and Success Criteria.} The robot must identify the single empty soda can among the four candidates and place it onto the designated wooden tray. At the beginning of the task, the agent is informed that any can with a mass of less than 100g is considered empty. A trial is scored as a \textit{Success} if the target empty can is placed completely within the geometric boundaries of the tray. A trial is marked as a \textit{Failure} if the robot places an incorrect can on the tray, or if it fails to select any can due to inaccurate mass estimation.

\textbf{Design Intent.} This task is designed to evaluate the agent's interactive perception and physical reasoning capabilities. First, the agent must use visual cues to deduce that only cans with straws are open, allowing it to logically prioritize those candidates for physical exploration. Second, it must physically interact with the prioritized cans to estimate their mass, successfully distinguishing the empty can from the partially full one based on the given 100g threshold.

\subsubsection{Task 2: Pick Ripe Avocado}

\begin{figure}[t]
  \centering
  \scalebox{0.95}{
  \includegraphics[
    width=1\linewidth ]{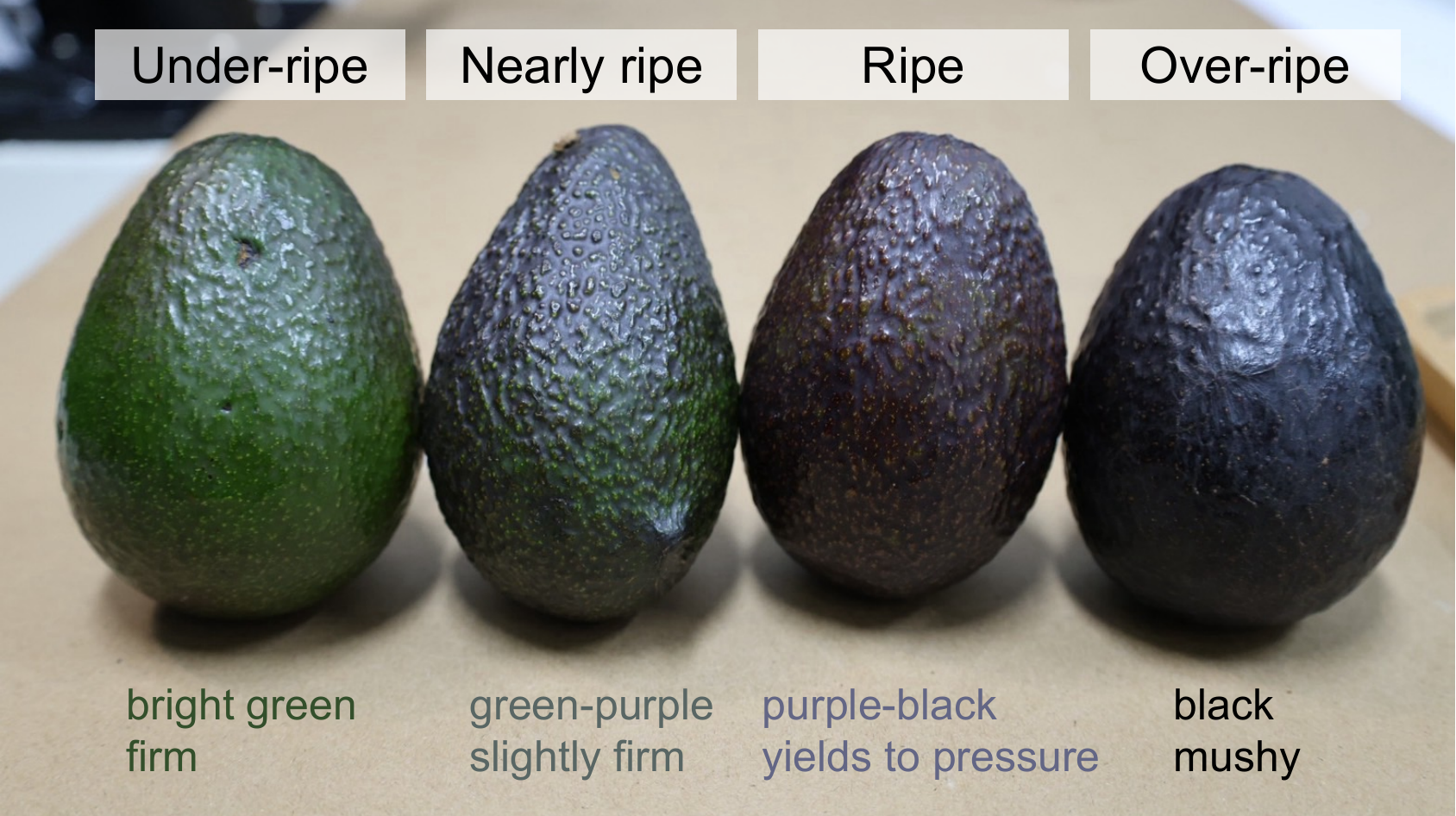}}
  \caption{\textbf{Visual progression of avocado ripening.} As the fruit matures, its skin transitions from bright green to black. This darkening serves as a primary visual cue to assess whether the avocado is ripe enough to consume.}
  \label{fig:item_avocado}
\end{figure}

\textbf{Scene Description.} The workspace contains four avocados, as illustrated in \Cref{fig:item_avocado}. Two of them have green skin, indicating they are visibly unripe, while the other two possess a darker skin color. Of the two dark avocados, one is ripe while the other is still hard. A wooden tray is provided for the robot to place the target avocado. 

\textbf{Objective and Success Criteria.} The agent must isolate the single perfectly ripe avocado from the group of four and transport it to the wooden tray. At the beginning of the task, the agent is informed that any avocado with a stiffness level of 2 or lower is considered ripe. A trial is scored as a \textit{Success} if the chosen avocado on the tray possesses a calibrated stiffness label corresponding to a ripe state ($s \le 2$). A trial is marked as a \textit{Failure} if the robot selects an unripe avocado (whether green or dark, $s \ge 3$), or if it fails to select any avocado due to inaccurate stiffness estimation.

\textbf{Design Intent.} This task is designed to evaluate the agent's ability to combine visual reasoning with interactive perception. First, the agent must use visual cues (see \Cref{fig:item_avocado}) to deduce that the darker avocados are more likely to be ripe, allowing it to logically prioritize those candidates for physical exploration. Second, it must physically interact with the prioritized avocados to estimate their stiffness, successfully identifying the truly ripe avocado based on the provided stiffness threshold.

\subsubsection{Task 3: Pack Grocery Bag}

\textbf{Scene Description.} The workspace contains four colored blocks representing grocery items and a grocery bag placed on the table. The blocks vary only in color, mass, and stiffness, while other factors such as volume, shape, and semantic object identity are held constant to isolate the effects of physical properties and reduce potential confounding factors.

\textbf{Objective and Success Criteria.}
The agent must identify a suitable foundation item for the grocery bag by considering both mass and stiffness. The scene is constructed with two lightweight items and two similarly heavy items. Among the two heavy candidates, one is substantially stiffer than the other. The agent should therefore first identify the heavier candidates through mass measurement and then distinguish between them using stiffness. A trial is scored as a success if the agent selects the stiffer of the two heavy candidates and places it at the bottom of the grocery bag.

\textbf{Design Intent.} This task evaluates whether the agent can accumulate and use multiple physical properties over a longer exploration horizon. Mass measurements first separate the two heavier candidates from the lightweight distractors. Because the two heavy candidates have similar mass, mass alone is
insufficient to determine the better foundation. The agent must therefore use stiffness measurements to resolve the remaining ambiguity and select the stiffer heavy candidate. This setup tests whether the agent can use previously acquired measurements to guide subsequent physical exploration rather than exhaustively measuring every property of every object.

\subsubsection{Performance on Vision-Only Tasks: Find Blue Cube}

\begin{wrapfigure}{r}{0.42\textwidth}
    \vspace{-0.4cm}
    \centering
    \includegraphics[width=0.58\linewidth]{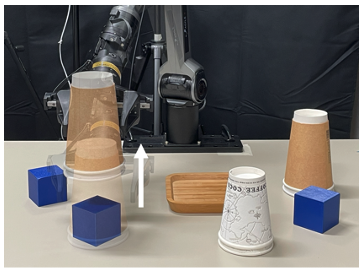}
    \hfill
    \includegraphics[width=0.36\linewidth]{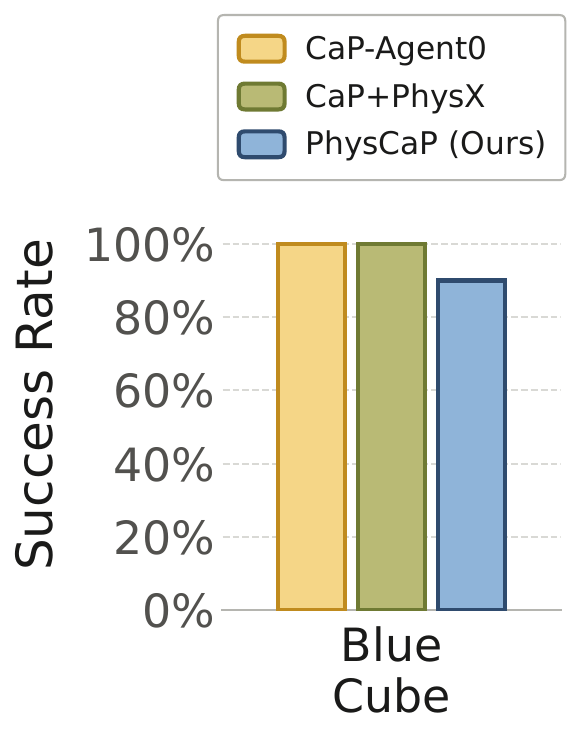}
    \caption{\textbf{Find Blue Cube task.} The cube is hidden under one of three cups, one of which is too small to hold it. All methods reach comparable success.}
    \label{fig:task-blue-cube}
    \vspace{-0.3cm}
\end{wrapfigure}

\textbf{Scene Description.} The workspace contains three paper cups and three uniformly sized blue cubes, two of which are initially visible. The third blue cube is concealed beneath one of the two larger paper cups. The third cup is intentionally too small to act as a container for a cube, as illustrated in \Cref{fig:task-blue-cube}. A wooden tray is provided for the robot to place explored cups, and a human assistant is present to help clear chosen cups from the workspace if necessary. 

\textbf{Objective and Success Criteria.} The objective is to locate and reveal the hidden blue cube so that all three cubes are simultaneously in sight. At the beginning of the task, the agent is explicitly informed that all three cubes share the same dimensions. A trial is scored as a \textit{Success} if the robot safely lifts the correct cup, leaving all three blue cubes fully exposed to the workspace camera view. A trial is marked as a \textit{Failure} if the robot knocks over a cup without properly lifting it, or if it terminates the sequence without revealing the hidden cube. 

\textbf{Design Intent.} This task is designed to evaluate whether the agent can utilize visual cues for physical reasoning. Specifically, the agent must perceive that the smallest cup is visually smaller than the exposed cubes, and logically deduce that it cannot contain the hidden cube. Consequently, a successful agent will demonstrate this physical reasoning by prioritizing the two larger cups for exploration.

\textbf{Does PhysCaP's capability to require physical measurement make it worse at solving tasks that can be solve with vision alone?}
To test this, we add a Find Blue Cube task in which the agent locates a blue cube hidden beneath one of three upside-down cups. Two identical cubes are left visible, so the agent can see the target's size but not its location, and some cups are too small to conceal a cube of that size. The task is solvable by geometric reasoning alone, with no measurement required. \Cref{fig:task-blue-cube} shows that PhysCaP performs on par with the vision-only baselines, confirming that the added physical reasoning does not interfere when it is unnecessary. The agent rules out the infeasible cups and lifts only the plausible ones, rather than measuring objects it has no reason to measure.

\subsection{Task-Specific Prompts}
\label{sec:app_exp_item}
\label{sec:app_task_prompts}
This section provides the exact task-specific prompts supplied to the agent for the three scenarios detailed in \Cref{sec:tasks}. These prompts describe what each task is about, how the task environment should be set up, the task objective, and some action restrictions for carrying out the task.

\textbf{Task 1: Identify Empty Can}
\begin{lstlisting}[
    language={},
    breaklines=true,
    breakatwhitespace=true,
    columns=fullflexible,
]
# Task: Find the empty can

## Environment
- A table surface containing four soft drink cans.
- One can is empty; the others are not.
- A target destination designated by a wooden tray.

## Objective
Systematically interact with the cans to identify the empty one and relocate it to the wooden tray.

## Operational Protocol
1. Observe & Anchor: Analyze the image. Identify all cans and assign IDs based on color and position (e.g., "Blue can left").
2. Reason: Output the reasoning for the proposed action.
   - Output Format: THOUGHT: [Reasoning] | ACTION: [Lift can ID].
3. Evaluate:
   - A can is classified as empty if the mass is below 0.1 kg.
   - If NOT empty: State "Can [ID] is not empty." Place it back at its original coordinates. Request next observation.
   - If empty: Announce "Empty can confirmed: [ID]." Proceed to pick and place the can onto the wooden tray.

## Constraints
- Sequential Interaction: Interact with exactly one object per reasoning cycle.
- State Tracking: Maintain a list of "Checked" vs. "Unchecked" IDs to prevent redundant loops.
- Termination: End the task only when the empty can is on the wooden tray or all cans are confirmed non-empty.

\end{lstlisting}

\textbf{Task 2: Pick Ripe Avocado}
\begin{lstlisting}[
    language={},
    breaklines=true,
    breakatwhitespace=true,
    columns=fullflexible,
]
# Task: Find the ripe avocado

## Environment
- A table surface containing four avocados.
- One avocado is ripe; the others are unripe.
- A target destination designated by a wooden tray.

## Objective
Systematically interact with the avocados to identify the ripe one and relocate it to the wooden tray.

## Operational Protocol
1. Observe & Anchor: Analyze the image. Identify all avocados and assign unique IDs based on color and position (e.g., "avocado_dark_left", "avocado_green_center").
2. Reason: Output the reasoning for the proposed action.
   - Output Format: THOUGHT: [Reasoning] | ACTION: [Gently squeeze Avocado ID].
3. Evaluate:
   - An avocado is considered ripe if it is soft (stiffness level <= 2).
   - If NOT ripe: State "Avocado [ID] is unripe." Leave it in its original position. Request next observation.
   - If ripe: Announce "Ripe avocado confirmed: [ID]." Proceed to pick and place the avocado onto the wooden tray.

## Constraints
- Sequential Interaction: Interact with exactly one object per reasoning cycle.
- State Tracking: Maintain a strict list of "Checked" vs. "Unchecked" IDs to prevent redundant loops.
- Termination: End the task only when the ripe avocado is successfully placed onto the wooden tray, or all avocados have been physically tested and confirmed unripe.
\end{lstlisting}

\textbf{Task 3: Pack Grocery Bag}
\begin{lstlisting}[
    language={},
    breaklines=true,
    breakatwhitespace=true,
    columns=fullflexible,
]
# Task: Pack the grocery bag

## Environment
- A table surface containing four coloured cubes and one paper grocery bag.
- The cubes are the groceries to be packed.

## Objective
You are packing the groceries into the paper bag. Items must be packed one at a
time, in a sensible order. Decide which single item should go into the bag
FIRST, put that item in the bag, and stop.

## Operational Protocol
1. Observe & Anchor: Analyze the image. Identify all cubes and assign IDs based on colour and position (e.g., "Blue cube left").
2. Reason: Output the reasoning thought for the proposed action.
   - Output Format: THOUGHT: [Reasoning] | ACTION: [Action on cube ID].
3. Commit: Once you have decided which item goes in first, print exactly one line
   in this format, on its own line:
     FIRST ITEM: [cube ID]
   Then place that cube in the paper bag.

## Constraints
- Sequential Interaction: Interact with exactly one cube per reasoning cycle.
- Single Placement: Put exactly one cube in the bag. Leave the other three on the table.
- Termination: End the task once the first item is in the paper bag.

\end{lstlisting}

\textbf{Vision-Only Task: Find Blue Cube}
\begin{lstlisting}[
    language={},
    breaklines=true,
    breakatwhitespace=true,
    columns=fullflexible,
]
# Task: Find all three blue cubes on the table.

## Environment
- There are exactly three blue cubes among a collection of objects on the table.
- All blue cubes are the same size and shape.
- Some blue cubes may be covered by other objects.
- Some cups may be too small to cover a blue cube.
- A wooden tray is also on the table.

## Objective
Systematically interact with the objects on the table so that all three blue cubes are in sight.

## Operational Protocol
1. Observe & Anchor: Analyze the image. Identify all objects on the table and assign IDs based on color and position (e.g., "Blue cube left").
2. Reason: Output the reasoning for the proposed action.
   - Output Format: THOUGHT: [Reasoning] | ACTION: [Lift cup ID].
3. Evaluate:
   - After lifting an object, check if any new blue cubes have come into view.
   - Place lifted objects on the wooden tray after checking. A human assistant will remove the object from the workspace.
   - If no new blue cubes are found: State "Did not find any new blue cubes." Place the interacted object on the wooden tray. Return to home pose and request the next observation.
   - Stop the task when all three blue cubes are in sight.

## Constraints
- Sequential Interaction: Interact with exactly one object per reasoning cycle.
- State Tracking: Maintain separate lists for "Located blue cubes" and "Checked objects" to prevent redundant loops.
- Position Format: Report positions as [x, y, z] in meters (robot base frame).
- Termination: End the task only when all three blue cubes are in sight.
\end{lstlisting}

\subsection{Qualitative Results}
\label{sec:app_code_demo}

As quantitatively shown in \Cref{tab:main-results}, the vision-only \textbf{CaP} baseline performs poorly on tasks that depend on hidden physical properties, as it has no mechanism to directly measure them. Adding the PhysX modules (\textbf{CaP+PhysX}) provides access to the required physical information, but without explicit exploration reasoning the agent tends to interact with objects exhaustively, resulting in unnecessary measurements and longer execution time. In contrast, \textbf{PhysCaP} combines a Planner, which determines when sufficient information has been collected, with a Prioritizer, which uses visual cues and previously acquired measurements to guide the order of subsequent interactions. This enables selective exploration: on the Empty Can and Ripe Avocado tasks, \textbf{PhysCaP} focuses on visually promising candidates, while on the longer-horizon Pack Grocery Bag task it uses accumulated physical measurements to progressively narrow the exploration space. Consequently, \textbf{PhysCaP} achieves the lowest or tied number of object interactions across all three tasks and the shortest execution time on two of the three tasks.

\subsubsection{Task 1: Identify Empty Can}

\begin{figure}[h]
    \centering
    \includegraphics[width=\linewidth]{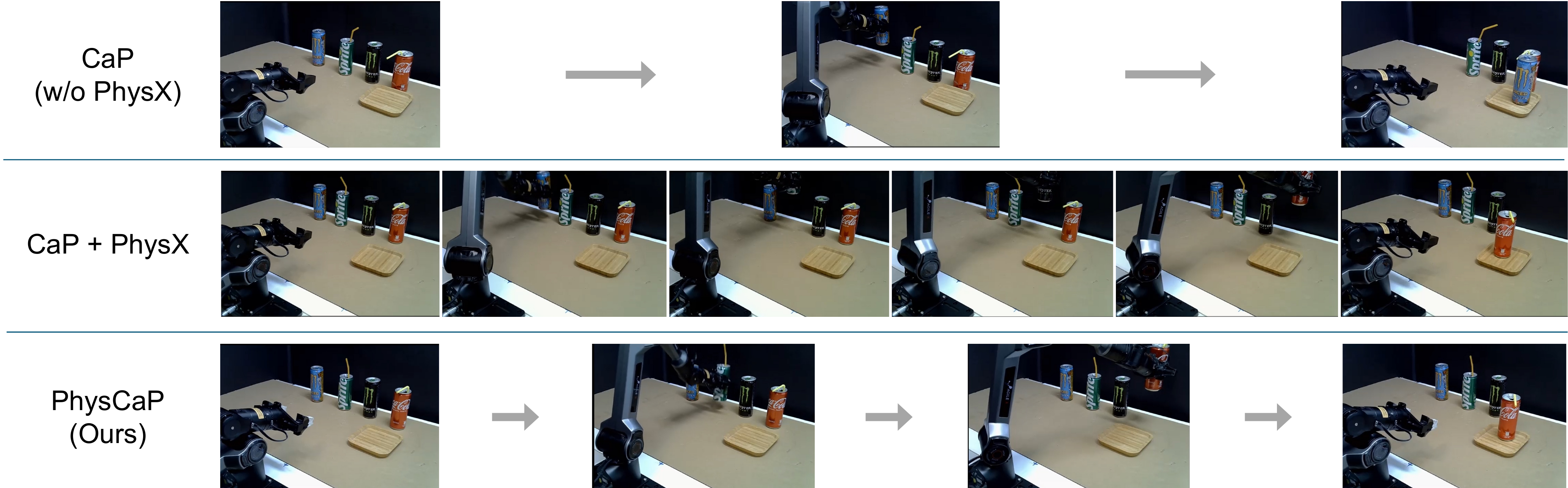}
    \caption{
    \textbf{Qualitative results for the Identify Empty Can task.}
    \textbf{CaP} drastically under-explores by blindly guessing which can is empty, as it inherently lacks the ability to infer hidden physical properties. The \textbf{CaP+PhysX} baseline over-explores by lifting and weighing every can in the scene. In contrast, our \textbf{PhysCaP} method efficiently weighs only the cans that present visual cues.
    }
    \label{fig:qual_can}
\end{figure}

\textbf{Trajectory Comparison.} The interaction step comparison is shown in \Cref{fig:qual_can}.
\begin{itemize}
\item \textit{CaP:} Fails completely due to its inability to obtain physical measurements. 
\item \textit{CaP+PhysX:} The agent exhaustively executes the \texttt{get\_mass} primitive on all four cans, sequentially lifting each one to read joint torques before making a decision. 
\item \textit{PhysCaP:} The Prioritizer maps the visual presence of a straw and an open tab to human behavioral semantics, inferring that sealed cans are full. It dynamically ranks the open cans at the front of the exploration queue. The Planner triggers a \texttt{get\_mass} check on the first open can; if it is empty, the exploration loop meets its stopping criterion, halts immediately, and proceeds to the final execution phase. 
\end{itemize}

\textbf{Code Policy Demonstration.} The code block below illustrates the PhysCaP generated policy for this task. The Prioritizer successfully leverages visual heuristics, specifically the presence of inserted straws, to deduce that the green and red cans are the most probable candidates, prioritizing them for physical exploration. Consequently, Code Blocks 0 and 1 command interactions with the green and red cans, respectively. Once the red Coca-Cola can is weighed and its mass confirms it is empty, the Planner's stopping criterion is satisfied. In the final code block, the agent's internal reasoning acknowledges that the target has been identified, executes the placement to the target location, and successfully terminates the procedure.

\begin{lstlisting}
# Code block 0
import numpy as np

print("THOUGHT: Checking the mass of the second from left green can to see if it is the empty one. | ACTION: Lift second from left green can")
mass = get_mass("second from left green can")
print(f"Mass of second from left green can: {mass}")

if mass < 0.1:
    print("Empty can confirmed: second from left green can")
    
    can_pos, can_quat = get_object_pose("second from left green can")
    tray_pos, tray_quat = get_object_pose("wooden tray")
    
    open_gripper()
    goto_pose(can_pos, can_quat, z_approach=0.1)
    close_gripper()
    
    # Move 10 cm above the object to avoid collisions
    approach_pos = can_pos.copy()
    approach_pos[2] += 0.1
    goto_pose(approach_pos, can_quat)
    
    # Calculate target position for placing
    target_pos = tray_pos.copy()
    target_pos[2] = can_pos[2] + 0.05
    
    # Move 10 cm above target
    target_approach = target_pos.copy()
    target_approach[2] += 0.1
    goto_pose(target_approach, can_quat)
    
    # Place on tray
    goto_pose(target_pos, can_quat)
    open_gripper()
    
    # Move above tray and object before going to home pose
    goto_pose(target_approach, can_quat)
    home_pose()
else:
    print("can second from left green can is not empty.")
    breakpoint_code_block()

# Code block 1
import numpy as np

print("THOUGHT: Checking the mass of the right-most red can to see if it is the empty one. | ACTION: Lift right-most red can")
mass = get_mass("right-most red can")
print(f"Mass of right-most red can: {mass}")

if mass < 0.1:
    print("Empty can confirmed: right-most red can")
    
    can_pos, can_quat = get_object_pose("right-most red can")
    tray_pos, tray_quat = get_object_pose("wooden tray")
    
    open_gripper()
    goto_pose(can_pos, can_quat, z_approach=0.1)
    close_gripper()
    
    # Move 10 cm above the object to avoid collisions
    approach_pos = can_pos.copy()
    approach_pos[2] += 0.1
    goto_pose(approach_pos, can_quat)
    
    # Calculate target position for placing
    target_pos = tray_pos.copy()
    target_pos[2] = can_pos[2] + 0.05
    
    # Move 10 cm above target
    target_approach = target_pos.copy()
    target_approach[2] += 0.1
    goto_pose(target_approach, can_quat)
    
    # Place on tray
    goto_pose(target_pos, can_quat)
    open_gripper()
    
    # Move above tray and object before going to home pose
    goto_pose(target_approach, can_quat)
    home_pose()
else:
    print("can right-most red can is not empty.")
    breakpoint_code_block()

# Code block 2
import numpy as np

print("THOUGHT: The empty can (right-most red can) has already been successfully identified and relocated to the wooden tray in the previous step. The task is fully complete and no further actions are necessary. | ACTION: End task")
home_pose()
\end{lstlisting}

\subsubsection{Task 2: Pick Ripe Avocado}
\begin{figure}[h]
    \centering
    \includegraphics[width=\linewidth]{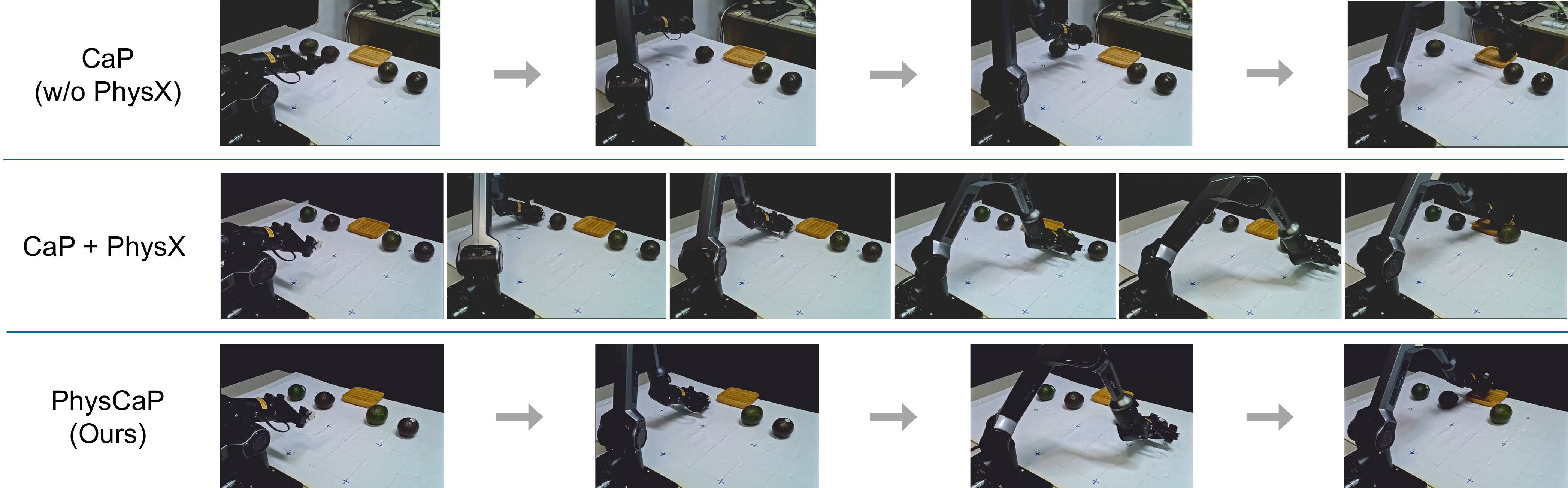}
    \caption{
    \textbf{Qualitative results for the Pick Ripe Avocado task.}
    The naive \textbf{CaP} baseline cannot detect hidden physical states and instead resorts to blind guessing, placing a random avocado on the tray. \textbf{CaP+PhysX} over-explores by squeezing every avocado before identifying the ripe one. In contrast, our \textbf{PhysCaP} method efficiently squeezes only the two visually dark avocados, finding the ripe one in the fewest steps.
    }
    \label{fig:qual_avo}
\end{figure}

\textbf{Trajectory Comparison.} The interaction step comparison is shown in \Cref{fig:qual_avo}.
\begin{itemize}
    \item \textit{CaP:} Fails completely due to its inability to obtain physical measurements. 
    \item \textit{CaP+PhysX:} The agent executes a tactile squeezing profile (\texttt{get\_stiffness}) across all four avocados indiscriminately, wasting significant operational time. 
    \item \textit{PhysCaP:} The Prioritizer uses color hints to isolate the two green avocados as explicitly unripe, removing them entirely from the execution graph. It prioritizes the two dark-skinned candidates. The robot executes the two-phase fine-stepping squeezing routine on the first dark avocado; if the inferred category returns as ripe, the Planner's stopping criterion is satisfied, avoiding any physical contact with the remaining objects.
\end{itemize}

\textbf{Code Policy Demonstration.} The PhysCaP generated policy highlights the agent's ability to correctly reason that the two dark-skinned avocados are the most promising candidates. Prioritizing stiffness measurements on these specific objects yields the most efficient strategy for task completion. As shown in Code Blocks 0 and 1, the agent interacts with the first dark avocado, and upon determining it is unripe, sequentially moves to test the second one. After successfully identifying the ripe avocado via stiffness estimation, the agent executes the required placement action. In the final code block, the agent confirms the ripe avocado has been successfully moved to the target location and terminates execution.

\begin{lstlisting}
# Code block 0
import numpy as np

print("THOUGHT: The current step is to test the dark avocado left of center (second from left). I will measure its stiffness to determine if it is ripe. If it is soft (stiffness <= 2), I will relocate it to the wooden tray. | ACTION: Gently squeeze avocado_dark_left")

avocado_name = "There are the fruits on the table are avocados, please point to the one that is the dark avocado second from the left."
stiffness = get_stiffness(avocado_name)
print(f"Stiffness of avocado_dark_left: {stiffness}")

if stiffness > 0 and stiffness <= 2:
    print("Ripe avocado confirmed: avocado_dark_left.")
    
    # Proceed to pick up the ripe avocado
    pos, quat = get_object_pose(avocado_name)
    open_gripper()
    goto_pose(pos, z_approach=0.1)
    close_gripper()
    
    # Move 10 cm above the object before going to the target position to avoid collisions
    goto_pose(pos + np.array([0.0, 0.0, 0.1]))
    
    # Get tray location
    tray_pos, tray_quat = get_object_pose("wooden tray")
    
    # Aim for the object's z-coordinate + 5 cm above the tray's surface
    target_pos = np.array([tray_pos[0], tray_pos[1], pos[2] + 0.05])
    goto_pose(target_pos, z_approach=0.1)
    open_gripper()
    
    # Move 10 cm above the tray and object before going to home pose
    goto_pose(target_pos + np.array([0.0, 0.0, 0.1]))
    home_pose()
else:
    print("Avocado avocado_dark_left is unripe.")
    # Request next observation
    breakpoint_code_block()

# Code block 1
import numpy as np

print("THOUGHT: The previous avocado was semi-rigid (unripe). The current step is to test the right-most dark avocado. I will measure its stiffness to determine if it is ripe. If soft (<= 2), I will relocate it to the wooden tray. | ACTION: Gently squeeze right-most dark avocado")

avocado_name = "There are the fruits on the table are avocados, please point to the one that is the right-most dark avocado."
stiffness = get_stiffness(avocado_name)
print(f"Stiffness of right-most dark avocado: {stiffness}")

if stiffness > 0 and stiffness <= 2:
    print("Ripe avocado confirmed: avocado_dark_right.")
    
    # Proceed to pick up the ripe avocado
    pos, quat = get_object_pose(avocado_name)
    open_gripper()
    goto_pose(pos, z_approach=0.1)
    close_gripper()
    
    # Move 10 cm above the object before going to the target position to avoid collisions
    goto_pose(pos + np.array([0.0, 0.0, 0.1]))
    
    # Get tray location
    tray_pos, tray_quat = get_object_pose("wooden tray")
    
    # Aim for the object's z-coordinate + 5 cm above the tray's surface
    target_pos = np.array([tray_pos[0], tray_pos[1], pos[2] + 0.05])
    goto_pose(target_pos, z_approach=0.1)
    open_gripper()
    
    # Move 10 cm above the tray and object before going to home pose
    goto_pose(target_pos + np.array([0.0, 0.0, 0.1]))
    home_pose()
else:
    print("Avocado avocado_dark_right is unripe.")
    # Request next observation
    breakpoint_code_block()

# Code block 2
print("THOUGHT: The right-most dark avocado was successfully identified as the ripe avocado (stiffness level 2) and has already been relocated to the wooden tray. The operational protocol has been fulfilled. | ACTION: Conclude task.")
\end{lstlisting}

\subsubsection{Task 3: Pack Grocery Bag}
\begin{figure}[h]
    \centering
    \includegraphics[width=\linewidth]{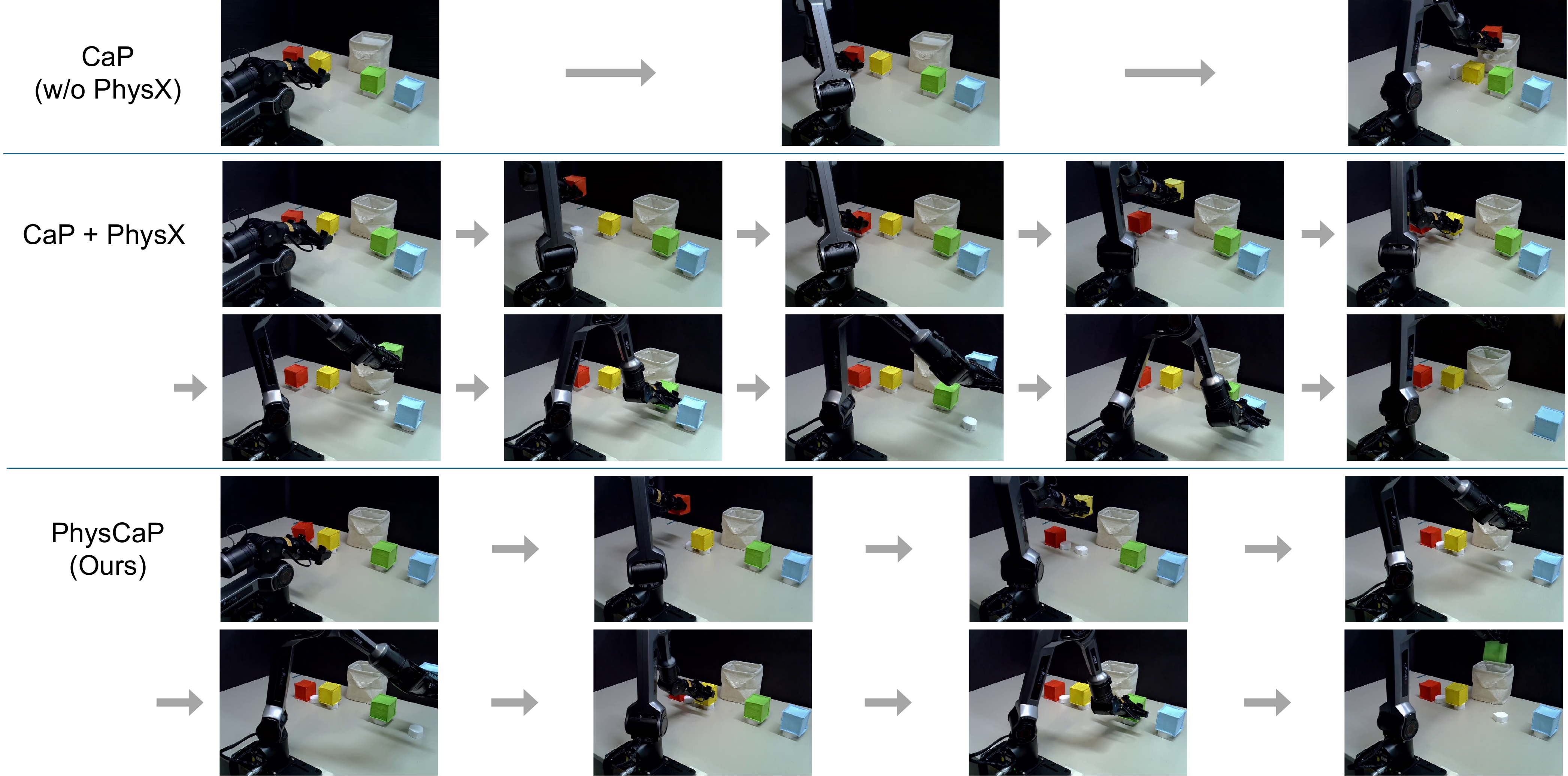}
    \caption{
    \textbf{Qualitative results for the Pack Grocery Bag task.}
    \textbf{CaP} cannot reliably select a suitable foundation item without access to hidden physical properties, while \textbf{CaP+PhysX} over-explores by measuring mass and stiffness across the candidate cubes. In contrast, \textbf{PhysCaP} uses accumulated mass measurements to identify the heavier candidates and prioritizes stiffness measurements on this reduced set, allowing it to select the stiffer heavy candidate with fewer unnecessary interactions.
    }
    \label{fig:qual_grocery}
\end{figure}

\textbf{Trajectory Comparison.} The interaction step comparison is shown in \Cref{fig:qual_grocery}.
\begin{itemize}
    \item \textit{CaP:} Fails to reliably select a suitable foundation item because the relative mass and stiffness of the objects cannot be determined from visual appearance alone.
    \item \textit{CaP+PhysX:} The robot alternates between using \texttt{{get\_mass}} and \texttt{{get\_stiffness}} for every cube. Without explicit prioritization or a sufficiency criterion, the agent therefore attempts to measure both the mass and stiffness of every cube. Moreover, over the longer task horizon, CaP+PhysX can lose track of previous measurements, leading to re-measurement or failure.
    \item \textit{PhysCaP:} The Planner keeps track of the physical properties measured during previous interactions, allowing the agent to use the accumulated information to guide subsequent exploration. After mass measurements reveal which cubes are heavier, the Prioritizer assigns higher priority to measuring the stiffness of these heavier candidates rather than continuing to explore all cubes indiscriminately. This allows PhysCaP to identify a cube that is both heavy and stiff earlier in the trajectory, reducing unnecessary measurements compared with CaP+PhysX.
\end{itemize}

\textbf{Code Policy Demonstration.}
The generated policy demonstrates how the Planner and Prioritizer progressively refine physical exploration using previously acquired measurements. The agent first measures the mass of all four cubes, allowing the Planner to retain their relative weights in the accumulated belief. Using this information, the Prioritizer narrows the subsequent stiffness exploration to the heavier yellow and green cubes, avoiding stiffness measurements on the lighter red and blue cubes. After measuring these two candidates, the agent identifies the green cube as one of the two heavy candidates and the more rigid of the two (stiffness level 5). The Planner then determines that sufficient information has been collected, terminates exploration, and generates the final policy to grasp the green cube and place it into the bag.
\begin{lstlisting}
# Code block 0
print("THOUGHT: I need to measure the mass of the red cube on the far left to determine its properties for packing order. Heavy and rigid items should be packed first at the bottom of the bag. | ACTION: Measure mass of cube_0")
mass = get_mass("the red cube on the far left")
print(f"Mass of cube_0: {mass:.3f} kg")
breakpoint_code_block()

# Code block 1
print("THOUGHT: I need to measure the mass of the yellow cube to help determine the packing order. | ACTION: Measure mass of cube_1")
mass_1 = get_mass("the yellow cube second from the left")
print(f"Mass of cube_1: {mass_1:.3f} kg")
breakpoint_code_block()

# Code block 2
print("THOUGHT: I need to measure the mass of the green cube to help determine the packing order. | ACTION: Measure mass of cube_2")
mass_2 = get_mass("the green cube second from the right")
print(f"Mass of cube_2: {mass_2:.3f} kg")
breakpoint_code_block()

# Code block 3
print("THOUGHT: I need to measure the mass of the blue cube to help determine the packing order. | ACTION: Measure mass of cube_3")
mass_3 = get_mass("the blue cube on the far right")
print(f"Mass of cube_3: {mass_3:.3f} kg")
breakpoint_code_block()

# Code block 4
print("THOUGHT: I need to measure the stiffness of the yellow cube to help determine if it should be packed at the bottom or top of the bag. | ACTION: Measure stiffness of cube_1")
stiffness_1 = get_stiffness("the yellow cube second from the left")
print(f"Stiffness of cube_1: {stiffness_1}")
breakpoint_code_block()

# Code block 5
print("THOUGHT: I need to measure the stiffness of the green cube to help determine if it should be packed at the bottom or top of the bag. | ACTION: Measure stiffness of cube_2")
stiffness_2 = get_stiffness("the green cube second from the right")
print(f"Stiffness of cube_2: {stiffness_2}")
breakpoint_code_block()

# Code block 6
import numpy as np

print("THOUGHT: Based on the measurements, the green cube is the heaviest (0.411 kg) and most rigid (level 5), making it the ideal foundational item for the bottom of the bag. | ACTION: Pack cube_2")
print("FIRST ITEM: cube_2")

# Get cube pose
cube_pos, cube_quat = get_object_pose("the green cube second from the right")

# Approach and grasp the green cube
open_gripper()
goto_pose(cube_pos, z_approach=0.1)
goto_pose(cube_pos)
close_gripper()
goto_pose(cube_pos, z_approach=0.1)

# Get bag pose
bag_pos, bag_quat = get_object_pose("the white paper bag in the middle")

# Aim for 5 cm above the bag's rim. 
# We add an offset to the bag's detected Z position to safely clear the rim.
drop_pos = bag_pos.copy()
drop_pos[2] += 0.15

# Approach the drop position (z_approach adds another 10 cm initially)
goto_pose(drop_pos, z_approach=0.1)

# Move down to exactly 5 cm above the bag's rim
goto_pose(drop_pos)

# Release the cube into the bag
open_gripper()

# Move 10 cm above the bag drop position before going home to avoid dragging it
goto_pose(drop_pos, z_approach=0.1)

# Return to rest pose
home_pose()
\end{lstlisting}

\subsubsection{Vision-only Task: Find Blue Cube}
\begin{figure}[h]
    \centering
    \includegraphics[width=\linewidth]{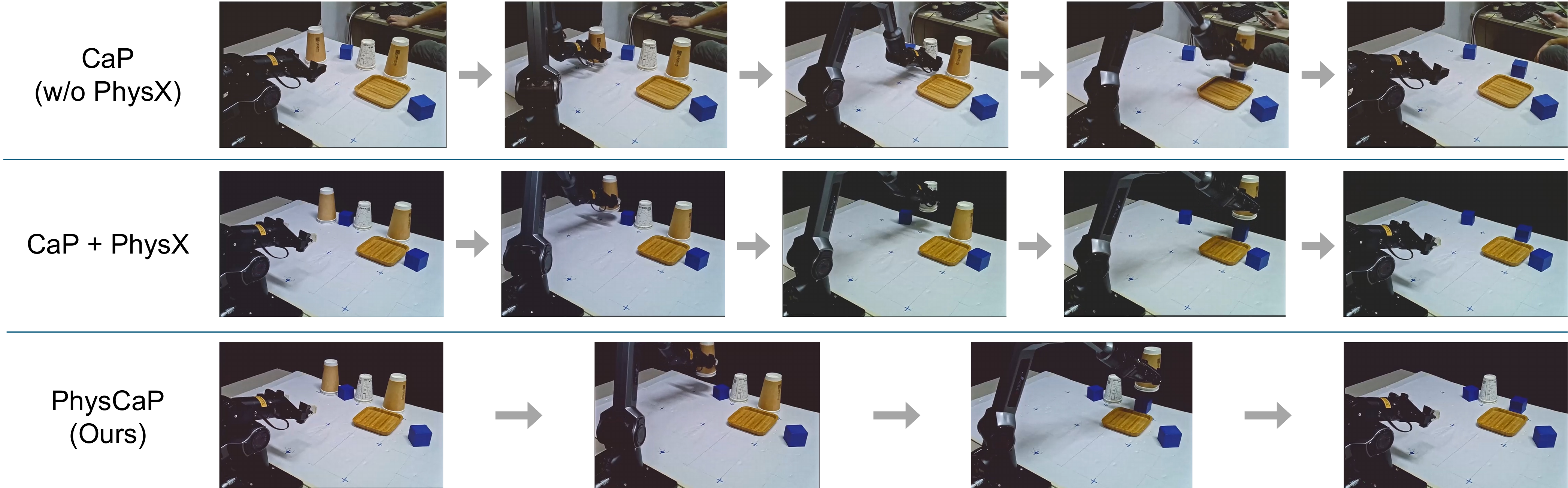}
    \caption{
    \textbf{Qualitative results for the Find Blue Cube task.}
    The \textbf{CaP} and \textbf{CaP+PhysX} methods exhaustively explore the scene by lifting all cups, whereas our \textbf{PhysCaP} method efficiently lifts only the visually hinted cups.
    }
    \label{fig:qual_cube}
\end{figure}

\textbf{Trajectory Comparison.} The interaction step comparison is shown in \Cref{fig:qual_cube}.
\begin{itemize}
    \item \textit{CaP/CaP+PhysX:} The agent lifts all three cups sequentially (including the obviously small coffee cup), consuming 3 interaction steps. 
    \item \textit{PhysCaP:} The Prioritizer utilizes geometric heuristics to immediately eliminate the small coffee cup from the action queue. The Planner then commands physical verification only on the remaining two viable cups, reducing the maximum interaction overhead to 1 or 2 steps depending on the cube's hidden location. 
\end{itemize}

\textbf{Code Policy Demonstration.} The following code demonstrates the PhysCaP generated policy for this task. In Code Block 0, the Prioritizer reasons that the first large coffee cup is a geometrically viable candidate that may conceal the target blue cube. Because this initial interaction reveals an empty space, the agent dynamically generates Code Block 1 to investigate the second large coffee cup. Lifting this second cup successfully reveals the hidden blue cube. Consequently, Code Block 2 terminates the sequence. The agent's internal reasoning confirms that all three target cubes are now visible in the workspace and officially marks the task as complete. 

\begin{lstlisting}
# Code block 0
import numpy as np

print("THOUGHT: To find the blue cubes, I will lift the left brown paper cup to check if a blue cube is underneath it. | ACTION: Lift left brown paper cup")

cup_pos, cup_quat = get_object_pose("left brown paper cup")
tray_pos, tray_quat = get_object_pose("wooden tray")

# Grasp the cup
open_gripper()
goto_pose(cup_pos, cup_quat, z_approach=0.1)
close_gripper()

# Lift the cup
goto_pose(cup_pos + np.array([0, 0, 0.1]), cup_quat)

# Calculate place position over the tray
place_pos = tray_pos.copy()
place_pos[2] = cup_pos[2] + 0.05

# Move to the tray and place
goto_pose(place_pos, tray_quat, z_approach=0.1)
open_gripper()

# Retreat
goto_pose(place_pos + np.array([0, 0, 0.1]), tray_quat)

# Return home
home_pose()
breakpoint_code_block()

# Code block 1
import numpy as np

print("THOUGHT: I will lift the right brown paper cup to check if another blue cube is hidden underneath it. | ACTION: Lift right brown paper cup")

cup_pos, cup_quat = get_object_pose("right brown paper cup")
tray_pos, tray_quat = get_object_pose("wooden tray")

# Grasp the cup
open_gripper()
goto_pose(cup_pos, cup_quat, z_approach=0.1)
close_gripper()

# Lift the cup
goto_pose(cup_pos + np.array([0, 0, 0.1]), cup_quat)

# Calculate place position over the tray
place_pos = tray_pos.copy()
place_pos[2] = cup_pos[2] + 0.05

# Move to the tray and place
goto_pose(place_pos, tray_quat, z_approach=0.1)
open_gripper()

# Retreat
goto_pose(place_pos + np.array([0, 0, 0.1]), tray_quat)

# Return home
home_pose()
breakpoint_code_block()

# Code block 2
print("THOUGHT: The third blue cube was revealed after lifting the right brown paper cup, meaning all three blue cubes are now in sight. The objective is achieved so the task is complete. | ACTION: None")
\end{lstlisting}

\subsection{Simulation}
\label{sec:app_exp_sim}
To compare our method with existing VLAs and ensure reproducibility, we replicated the Identify Empty Can task in the LIBERO environment. 

\subsubsection{Environment Design Details}
We designed the environment using LIBERO's tabletop setup, where the agent controls a simulated 6-DOF Franka arm. The workspace contains four colored cups and a target basket. To ensure reliable object recognition within the simulation, we replaced the soda cans with simulated cups of hand-shaken beverages. To match the visual semantics of the real-world task, the cups are modeled with a height of $8.27 \text{cm}$ (approximately $10.74 \text{cm}$ when an inserted straw is present). The primary collision cylinder for each cup possesses a diameter of $5.6 \text{cm}$ and a height of $10.0 \text{cm}$. To simulate the physical properties, the empty white cup is assigned a mass of $0.035 \text{kg}$, while the other full cups (red, yellow, and orange) are assigned a mass of $0.350 \text{kg}$. The basket asset was sourced from the standard LIBERO library. For each trial, the initial positions of the four cups are randomly sampled from seven pre-defined rectangular spawn regions on the table, with a slight spatial perturbation applied afterward.
\begin{figure}[t]
  \centering
  \scalebox{0.95}{
  \includegraphics[
    width=1\linewidth ]{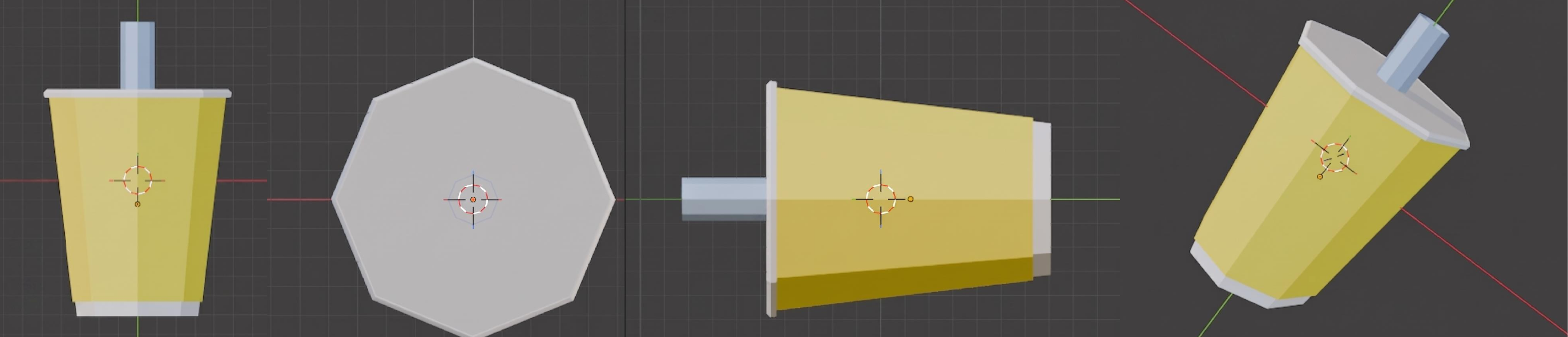}}
  \caption{A quad-view rendering of the custom 3D cup asset utilized in the simulated experiments. The completely opaque design is engineered to reliably conceal target objects, necessitating the interactive exploration behaviors evaluated in our benchmarking tasks.}
  \label{fig:asset_cup}
\end{figure}

\subsubsection{Task Details and Success Criteria} 

Mirroring the real-world Identify Empty Can task (detailed in \Cref{sec:app_task_can}), the agent is tasked with locating the empty cup and placing it into the basket. At the beginning of the task, the agent is explicitly informed that any cup with a mass of less than $100 \text{g}$ is considered empty. A trial is scored as a \textit{Success} if the empty white cup is placed entirely inside the basket. A trial is marked as a \textit{Failure} if these requirements are not met within 4000 time steps. 

\subsubsection{Task-Specific Prompts}
\begin{lstlisting}[
    language={},
    breaklines=true,
    breakatwhitespace=true,
    columns=fullflexible,
]
You are controlling a Franka Emika robot with privileged LIBERO APIs.

# Environment
- A table surface containing four colored cups: red, yellow, white, and orange.
- One cup is empty; the others are not.
- A target destination designated by a basket.

# Object Labels  (use these as stable IDs - do not invent names)
- red_cup_1 / "red_cup": a red cup
- yellow_cup_1 / "yellow_cup": a yellow cup
- white_cup_1 / "white_cup": a white cup
- orange_cup_1 / "orange_cup": an orange cup
- basket_1  / "basket": target basket
Straw visibility is an anchoring feature for mapping labels to objects and a weak task-specific exploration cue: a cup with a straw may have been used/opened. Use this cue for probe order only when candidates are otherwise comparable, and still require mass sensing before placement.
Coordinate grounding for this camera: larger x appears closer/front/lower in the image; smaller x appears farther/back/upper. Smaller y appears on the left in the image; larger y appears on the right in the image.

# Objective
Systematically probe cups one at a time via mass sensing to find the empty one, then place it in the basket.

# API Reference  (all pre-imported; import numpy explicitly if needed)
| Function | Returns |
|---|---|
| get_all_object_poses() | dict[label_str -> (pos_array, quat_array)] |
| get_object_pose(name: str) | (pos_array, quat_array) |
| sample_grasp_pose(name: str) | (pos_array, quat_array) |
| goto_pose(pos, quat, z_approach=0.10) | None; z_approach is pre-lift height in meters |
| goto_home_joint_position() | None; returns the arm to the saved default/reset joint position |
| close_gripper() / open_gripper() | None |
| get_observation() | obs dict |

Key observation fields:
- obs["object_masses"]: dict[str, float | None]
  keys are full LIBERO object IDs ("red_cup_1", "yellow_cup_1", "white_cup_1", "orange_cup_1", "basket_1")
  None -> not yet measured; float -> mass in kg after lifting
  For a short API label like "white_cup", read the corresponding mass from "white_cup_1".

Mass rule: a measured mass < 0.1 kg means the cup is empty.
A mass value of None means the probe did not reveal the mass yet; do not conclude the cup is not empty from None.

# Operational Protocol
1. Observe & Anchor: Call get_all_object_poses() and print all poses to confirm label-to-position mapping.
2. Probe: Lift one unmeasured cup at a time and print obs["object_masses"]. Prefer unmeasured straw cups before sealed/no-straw cups only when accumulated mass knowledge, grounding clarity, and action cost are otherwise comparable.
3. Evaluate:
- If NOT empty: State "Cup [ID] is not empty." Place it back at its original coordinates, release it, call goto_home_joint_position(), then request the next observation.
- If empty: Announce "Empty cup confirmed: [ID]." Place it back or keep it securely grasped only if immediately proceeding to basket placement; after the probe interaction is finished, call goto_home_joint_position() before planning the next subtask unless doing so would drop a grasped cup.
4. Place: Pick the confirmed empty cup and put it into the basket. After releasing the cup in the basket, call goto_home_joint_position() and then check get_observation().


# Constraints
- Sequential Interaction: Lift exactly one cup per reasoning cycle.
- If choosing among unmeasured cups, do not use label list order or alphabetical order as a priority signal. Use the weak straw-as-used/opened cue, visual grounding, printed poses, and action cost.
- Termination: End the task only when the empty cup is in the basket or all cups are confirmed non-empty.

# Key rules:
- Write code to interact with the environment.
- You will also receive visual feedback of the table. Examine the image.
- IMPORTANT: To remember the value for your next step, you MUST explicitly `print()` the result so you can read it in the console stdout.
- Move in closed loop and check get_observation() after probing or placing.
- Do not place a cup in the basket until its mass has been revealed and compared against the other cups.
- Avoid collisions by using vertical clearance: when moving a grasped cup, first lift or move to at least 10 cm above the cup/table before translating toward the target position, then descend only after the gripper is above the target.
- When placing the empty cup in the basket, keep the cup at least 10 cm above the basket while moving over it, then lower into the basket instead of dragging directly across the table height.
- Arm default posture: after every completed probe, failed probe recovery, non-empty cup return, confirmed placement, or other finished subtask, retreat vertically if holding or near an object, release only when appropriate, then call goto_home_joint_position() before the next reasoning/action cycle. Do not call goto_home_joint_position() while carrying a cup unless you have already lifted it safely and the path will not collide.


Write ONLY executable Python code (no code fences). If you want to use numpy, import it explicitly.
\end{lstlisting}

\subsubsection{VLA Model Checkpoints} 

To compare PhysCaP against state-of-the-art VLAs, including OpenVLA~\cite{kim2024openvla}, $\pi_{0.5}$~\cite{intelligence2025pi_}, and MolmoAct2~\cite{fang2026molmoact2}, we utilized their publicly released LIBERO checkpoints from the following sources:
\begin{itemize}
\item OpenVLA: \href{https://huggingface.co/openvla/openvla-7b-finetuned-libero-object/tree/main}{openvla/openvla-7b-finetuned-libero-object}
\item $\pi_{0.5}$: \href{https://huggingface.co/lerobot/pi05_libero_finetuned_v044}{lerobot/pi05\_libero\_finetuned\_v044}
\item MolmoAct2: \href{https://huggingface.co/allenai/MolmoAct2-LIBERO}{allenai/MolmoAct2-LIBERO}
\end{itemize}

\subsubsection{Qualitative Results}
\begin{figure}
    \centering
    \includegraphics[width=\linewidth]{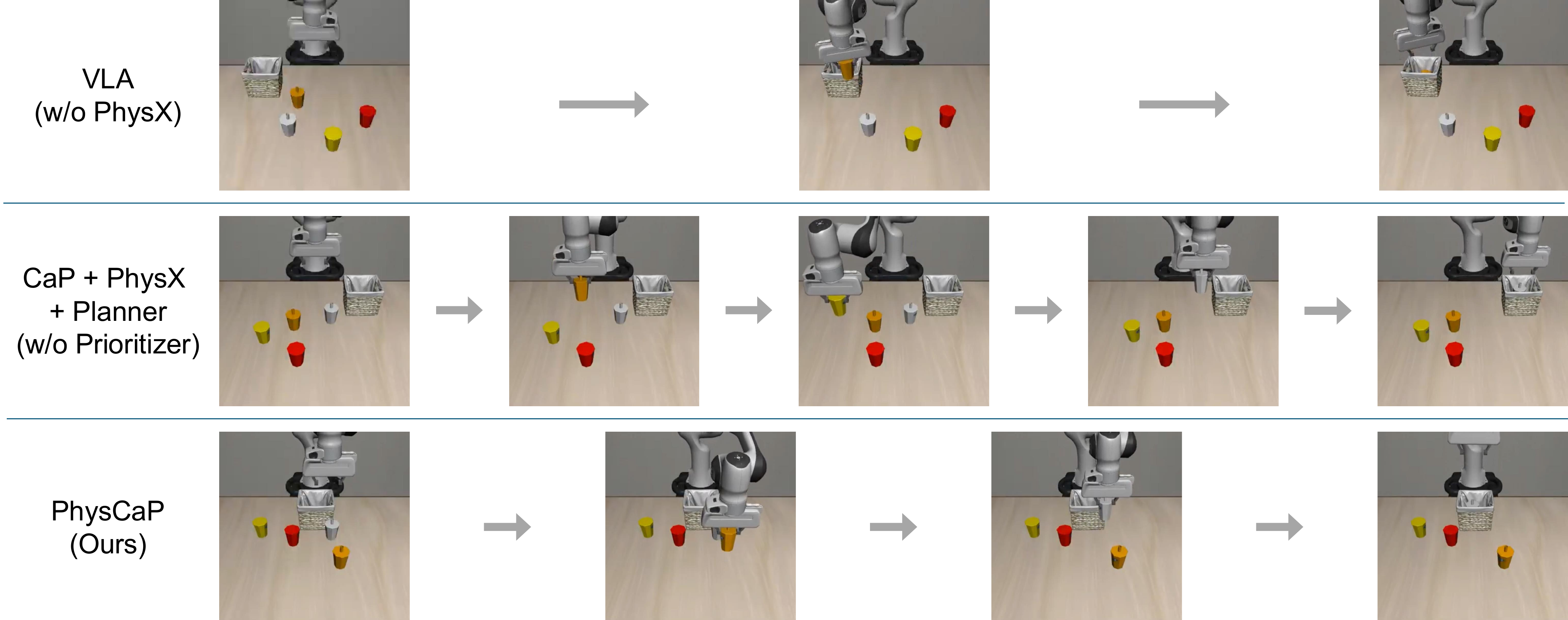}
    \caption{
    \textbf{Qualitative results for the Identify Empty Can task in the LIBERO environment.}
    The results demonstrate that the VLA baseline guesses blindly without physical feedback, while \textbf{CaP+PhysX+Planner} exhaustively over-explores by weighing every container. In contrast, our \textbf{PhysCaP} method efficiently targets and interactively weighs only the most probable candidates, completing the objective with superior accuracy and speed.
    }
    \label{fig:qual_sim}
\end{figure}

As shown in \Cref{fig:qual_sim}, even advanced VLAs like MolmoAct2 succeed only in picking up random objects due to their inability to actively measure mass. In contrast, PhysCaP not only successfully identifies the correct cup through active mass measurement, but also accomplishes the task with high efficiency by logically prioritizing the candidates for exploration.

\subsection{Item Specifications}
\label{sec:app_item_spec}

The tabletop objects used in our experiments comprise both standardized commercial items and open-source 3D-printed parts to ensure exact experimental reproducibility. The detailed specifications, geometric dimensions, and sourcing for these experimental props are outlined below.

\subsubsection{Task 1: Identify Empty Can}
\begin{figure}[t]
  \centering
  \scalebox{0.95}{
  \includegraphics[
    width=1\linewidth ]{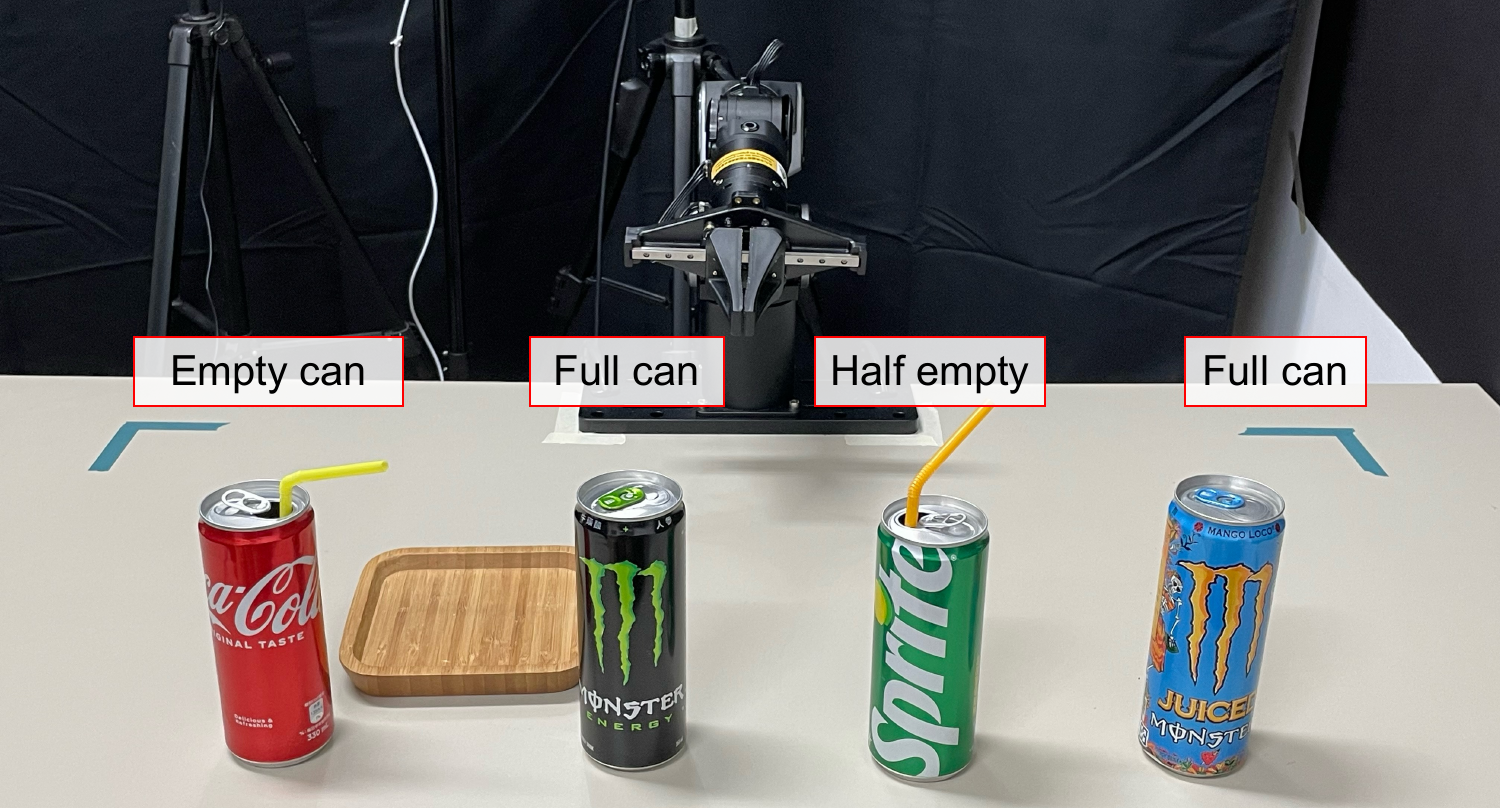}}
  \caption{
  \textbf{The object assets used in the Identify Empty Can task.} 
  The assets include two full cans, one half-full can, and the target empty can. This intentional variation in mass requires the robotic agent to interactively weigh the visually hinted containers to successfully identify the target.}
  \label{fig:item_empty_can}
\end{figure}

This task utilizes standard \href{https://a.co/d/0e8pgflL}{12~fl~oz soda cans} in three distinct physical states, as shown in \Cref{fig:item_empty_can}, to evaluate the agent's semantic and physical reasoning capabilities.

\textbf{Full cans:} Two unopened, factory-sealed cans are placed in the workspace. Because they are sealed, they visually imply a full state. These serve as semantic distractors that an efficient agent should eliminate from its physical exploration sequence.

\textbf{Half-full can:} This is a single opened can with a plastic straw inserted. To safely simulate a partially consumed beverage without the risk of liquid spills during robotic manipulation, the liquid was emptied and replaced with approximately $208\text{g}$ of metallic coins.

\textbf{Empty can:} The target object is a single opened can, also featuring an inserted straw, with all contents completely removed. This can weighs approximately $15\text{g}$, securely falling below the $100\text{g}$ empty threshold criteria.

\subsubsection{Task 2: Pick Ripe Avocado}
\begin{figure}[t]
  \centering
  \scalebox{0.95}{
  \includegraphics[
    width=1\linewidth ]{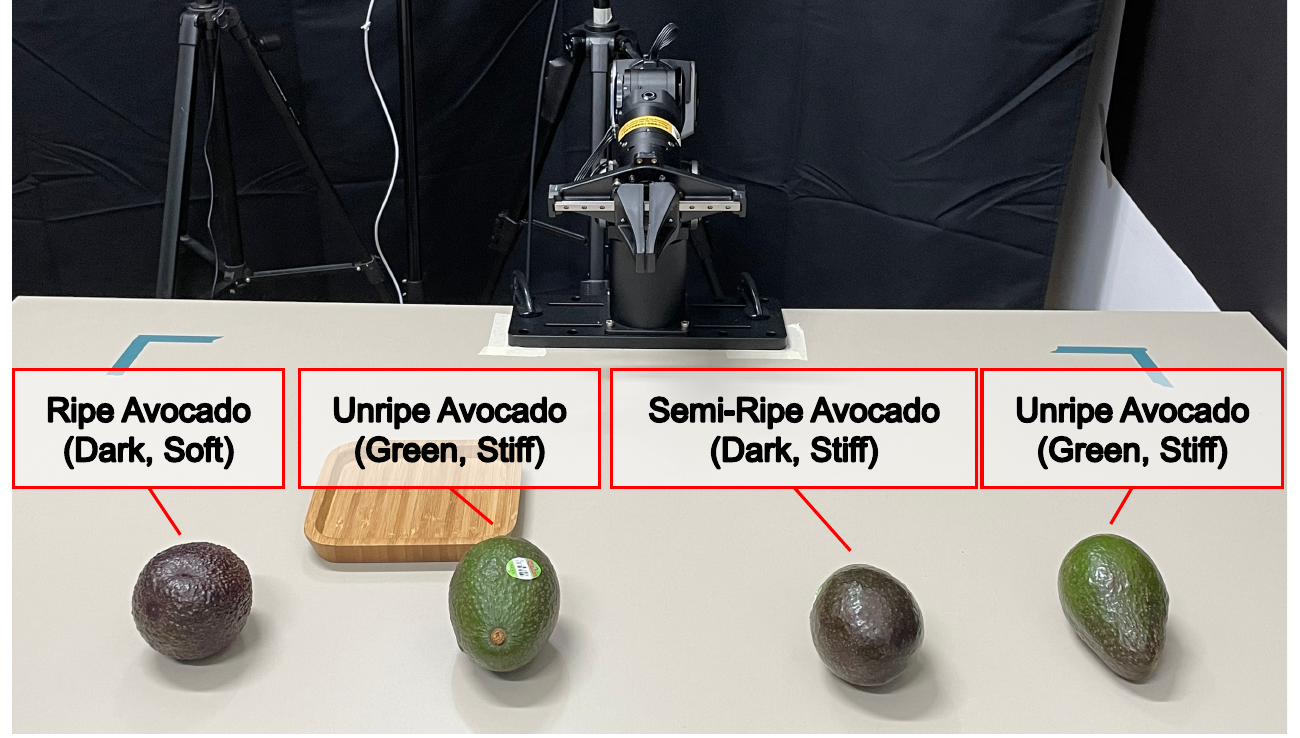}}
  \caption{  
  \textbf{The object assets used in the Pick Ripe Avocado task.} 
  The assets include two green avocados that are unripe and stiff, one dark avocado that's not ripe enough and still a bit stiff, and another dark avocado that's both ripe and soft. This intentional variation in stiffness requires the robotic agent to interactively test the stiffness of the visually hinted fruits to successfully identify the target.}
  \label{fig:item_avocado_2}
\end{figure}

This task utilizes standard \href{https://a.co/d/09gbBGOi}{Hass avocados} representing three distinct stages of ripeness, as illustrated in \Cref{fig:item_avocado_2}. This setup evaluates the agent's ability to synthesize visual heuristics with physical stiffness verification.

\textbf{Green unripe avocados:} These avocados possess a light green exterior, visually indicating an unripe state. They serve as semantic distractors that an efficient agent should immediately remove from its physical exploration sequence without requiring tactile interaction.

\textbf{Black unripe avocado:} A single dark-skinned, firm avocado is included in the workspace. Because its color visually mimics maturity, the agent cannot rely on visual priors alone. Instead, it must actively measure the object's rigidity to correctly eliminate it as a viable candidate.

\textbf{Black ripe avocado:} The target object is a single dark-skinned, ripe avocado that yields to gentle pressure. The agent must successfully identify and select this specific avocado to complete the task.

\subsubsection{Task 3: Pack Grocery Bag}

\begin{figure}[t]
  \centering
  \scalebox{0.95}{
  \includegraphics[
    width=1\linewidth ]{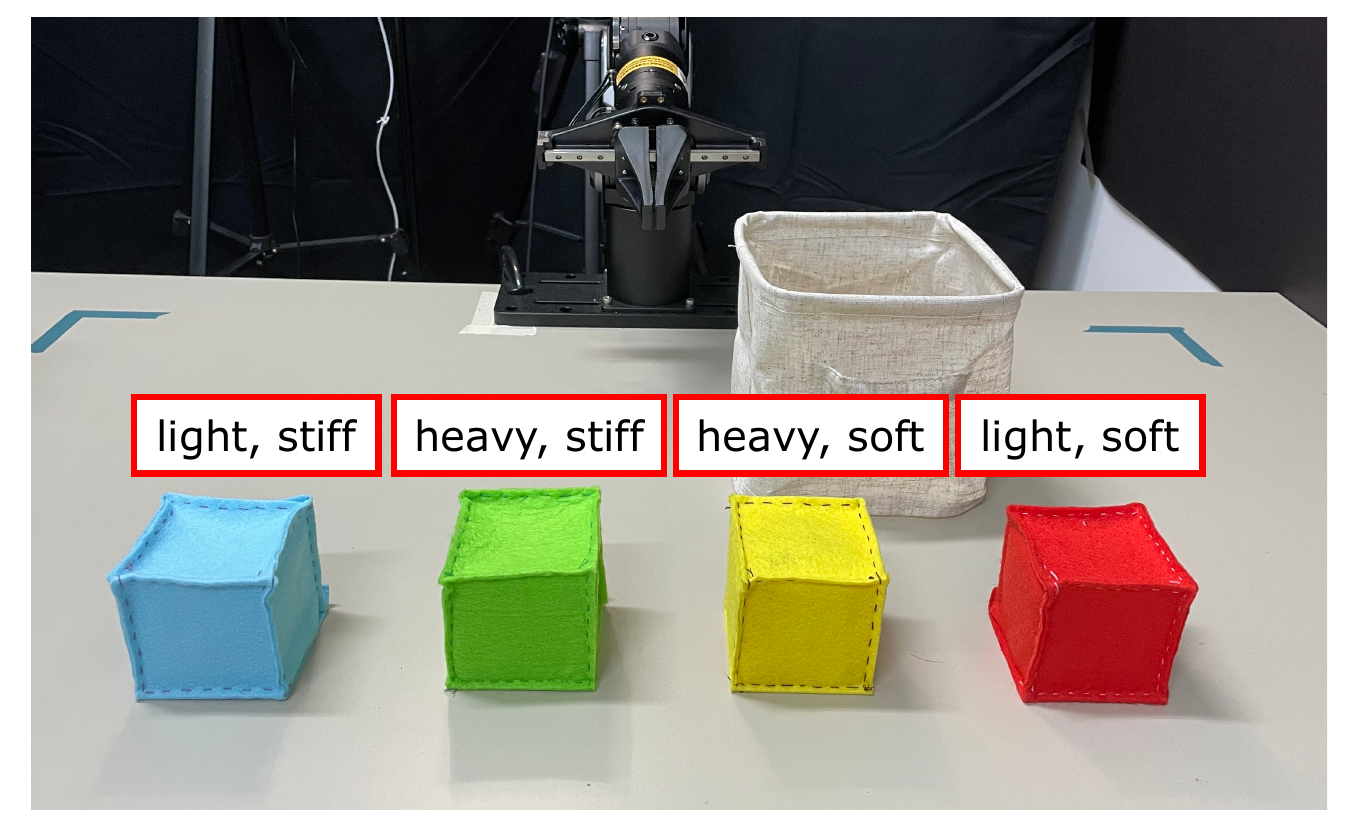}}
  \caption{
  \textbf{The object assets used in the Pack Grocery Bag task.} 
  The assets include four colored boxes with varying levels of stiffness and mass. These variations require the robotic agent to interactively weigh and squeeze the blocks to identify the target successfully.}
  \label{fig:item_grocery}
\end{figure}

This task uses four colored cubes to represent grocery items placed into a grocery bag, as shown in \Cref{fig:item_grocery}. Each cube is characterized by a different combination of mass and stiffness, allowing us to isolate physical properties that influence packing order. The agent must identify the item that provides the most stable foundation through physical mass and stiffness measurements.

\textbf{Blue cube.} This cube is light and stiff, representing packaged goods such as cereal boxes. Although relatively rigid, its low mass makes it more suitable for placement above heavier items to reduce the risk of bag tilting.

\textbf{Red cube.} This cube is light and soft, representing fragile goods such as bread or bags of chips. These items should be placed near the top of the bag to minimize compression from heavier objects.

\textbf{Yellow cube.} This cube is heavy and soft, representing dense but deformable goods such as bags of flour or ground meat. These items are better placed in the middle of the stack to avoid excessive compression while preventing them from crushing lighter items.

\textbf{Green cube.} This cube is heavy and stiff, representing structurally rigid goods such as canned food or milk cartons. It is the target object and provides the most stable foundation for the bottom of the grocery bag.

\subsubsection{Vision-Only Task: Find Blue Cube}

\begin{figure}[t]
  \centering
  \scalebox{0.95}{
  \includegraphics[
    width=1\linewidth ]{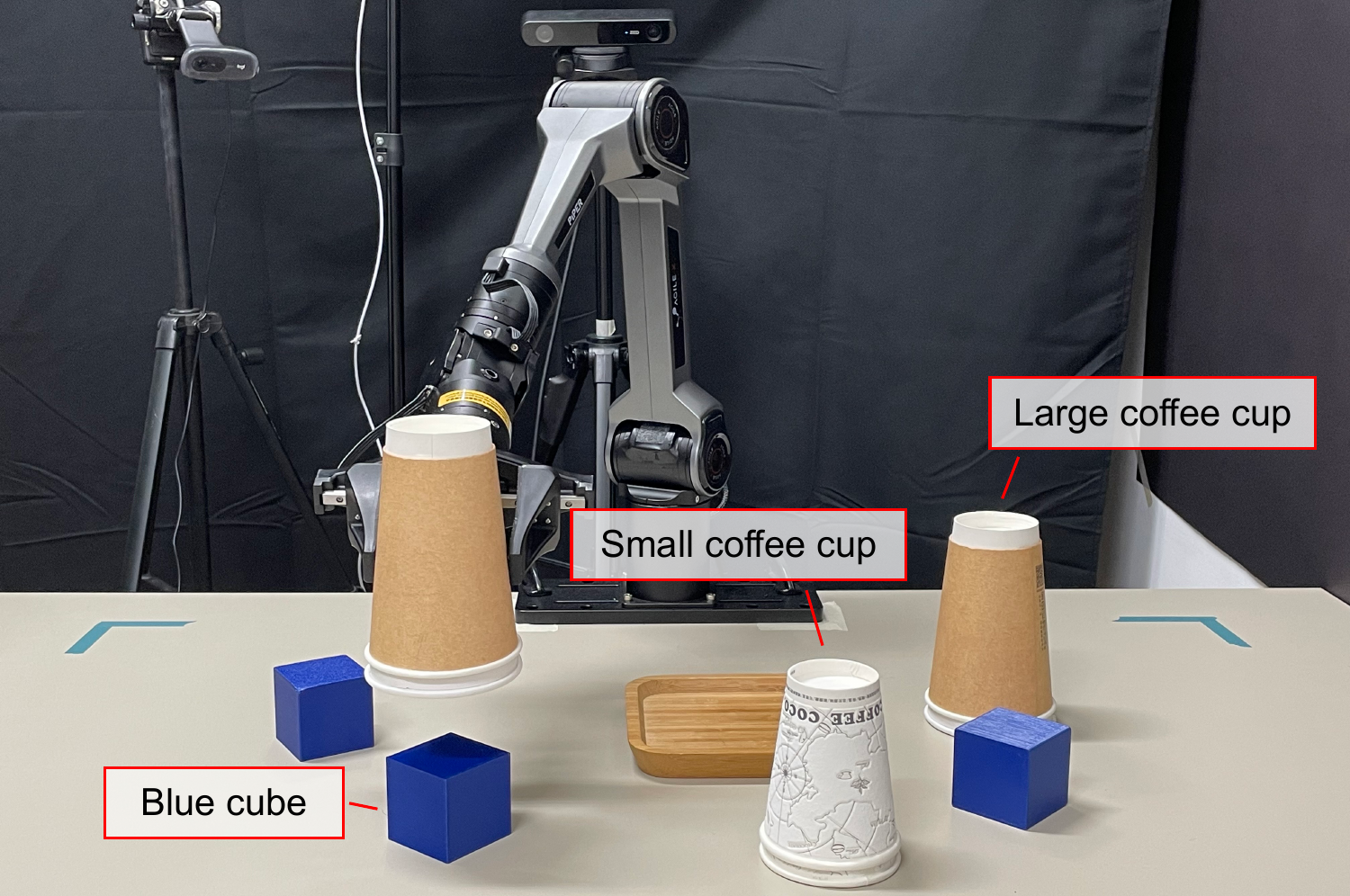}}
  \caption{
  \textbf{The object assets used in the Find Blue Cube task.} 
  The assets include the target blue cubes, large coffee cups designed to conceal the target, and a smaller cup acting as a visual distractor. This specific arrangement forces the robotic agent to employ interactive perception to successfully locate the hidden item among the containers.}
  \label{fig:item_blue_cube}
\end{figure}

This task utilizes 3D-printed blue cubes and a variety of coffee cups to evaluate the agent's visual reasoning capabilities, as illustrated in \Cref{fig:item_blue_cube}.

\textbf{Blue cubes:} Three identical blue cubes ($5\text{cm} \times 5\text{cm} \times 5\text{cm}$) serve as the target objects. They are placed in the workspace during task initialization and can be exactly replicated by 3D printing the provided \texttt{blue\_cubes.stl} file. 

\textbf{Large coffee cups:} These act as viable concealment containers, as their internal volume is geometrically sufficient to fully cover a blue cube when placed upside down. To recreate this scene, standard \href{https://a.co/d/0iAHJZzd}{16oz coffee cups} can be purchased online.

\textbf{Small coffee cup:} This object serves as a visual distractor. Because it is geometrically too small to conceal a blue cube, it tests the model's ability to utilize geometric reasoning to eliminate unviable candidates from its physical exploration sequence. Standard \href{https://a.co/d/0ibXVc07}{12oz coffee cups} can be purchased online to fulfill this role.

\section{Physical Property Extraction Module Details}
\label{sec:app_physx}

\subsection{Mass Measurement}

The \texttt{get\_mass} operation extracts an object's mass entirely from internal joint torques, eliminating the need for dedicated force sensors. The procedure begins by isolating the target object via the \texttt{get\_object\_pose()} API. Then the Molmo~2 model identifies the object within the global 2D camera view, and the ZED depth map back-projects this pixel into a precise 3D world coordinate. Guided by this spatial target, the robot arm navigates to a predefined lift pose 15~cm directly above the object. After a brief pause to allow kinematic vibrations to settle, the system records a baseline joint torque vector ($\tau_{\text{empty}}$) that captures the gravitational load of the bare arm. The arm then descends, grasps the object, and returns to the identical lift pose. Following another settling period, it records the loaded torque vector ($\tau_{\text{loaded}}$). Once the measurement is complete, the object is returned to its original position, and the arm resets to a home configuration. To compute the mass, the system calculates the differential torque ($\Delta\tau = \tau_{\text{loaded}} - \tau_{\text{empty}}$), isolating the object's physical load after applying hardware-specific gain corrections. Grounded in the principle of virtual work, the system uses the z-component of the arm's linear Jacobian ($\mathrm{J}_z$) to evaluate each joint's specific moment arm against gravity. Finally, the mass is estimated by projecting the torque differential onto this Jacobian and dividing by gravitational acceleration ($g = 9.8\text{ m/s}^2$), yielding a pose-invariant measurement that remains mathematically consistent regardless of the object's location in the workspace.

\textbf{Evaluation.} To evaluate the physical accuracy and robustness of the \texttt{get\_mass} module, we benchmarked its performance by repeatedly weighing five reference calibration masses spanning a broad range from 13g to 963g. Each object was measured across 50 independent trials to capture and account for real-world mechanical variability. Relying purely on the PiPER arm's internal motor current and proprioceptive joint torque feedback, the system successfully inferred both absolute and relative mass profiles. As illustrated by the resulting distributions in \Cref{fig:phys-mass}, the module maintains strict stability in capturing relative mass differences, with a slight increase in variance observed only at the extremes of the mass spectrum. Ultimately, these quantitative results demonstrate that the module provides a highly reliable physical prior for downstream reasoning tasks, such as cleanly distinguishing between empty and full containers.

\subsection{Stiffness Measurement}
\label{sec:app_details_stiff}
\label{sec:app_physx_eval}

The \texttt{get\_stiffness} operation quantifies object rigidity through a controlled, two-phase tactile squeezing procedure. To ensure measurement consistency, every call begins by homing the arm and fully opening the gripper. The system uses the established visual perception pipeline, leveraging Molmo~2 \cite{clark2026molmo2} for 2D semantic pointing and ZED depth maps for 3D back-projection, to navigate the gripper to the target object's spatial coordinates. Once positioned, the first phase establishes true physical contact. Because continuous gripper closing can generate internal mechanical friction that mimics object resistance, the jaws close in fine increments and perform a brief ``backoff'' verification test when the estimated force exceeds a threshold. By slightly reopening the jaws and verifying a corresponding decrease in force ($\Delta F$), the system distinguishes genuine object contact from internal friction. The verified contact position is then recorded as the reference displacement $d_0$.

In the second phase, the gripper resumes closing while recording discrete force-displacement pairs $(d_i, F_i)$ until reaching a predefined target force $F^*$. The contact segment is identified from these measurements, and the stiffness coefficient is estimated by fitting a linear model using ordinary least squares

\begin{equation}
k=\frac{\Delta F}{\Delta d} \mathrm{~(N/m)}.
\end{equation}

A larger slope indicates that force increases more rapidly with displacement and therefore corresponds to a stiffer object. The resulting stiffness coefficient is mapped to a discrete stiffness level $s \in \{1,\dots,5\}$, ranging from Ultra-Soft (1) to Rigid (5), using pre-calibrated boundaries. To mitigate measurement variance, each object is evaluated five times per session, and the final stiffness level is determined by majority vote.

\textbf{Evaluation.}
The physical intuition of this measurement technique is validated through a two-step evaluation process. To establish a reproducible ground truth for the stiffness evaluations, we designed a custom 3D-printed button mechanism. This device features a linear track with interchangeable springs that provide a known and reproducible resistance. Each spring has a stiffness of $k=420$ N/m, and placing multiple springs in parallel increases the effective stiffness in discrete increments. We use four calibrated boundaries to partition the measurement range into five stiffness levels, from Ultra-Soft to Rigid. These levels are not standardized material grades but reference points selected to span the stiffness range of the task objects, and the spring-based design allows the calibration to be readily rebuilt and reproduced.

To validate the metric on real objects, we evaluate three avocados with different ripeness levels using repeated squeezing trials while recording jaw displacement and estimated gripper force throughout each interaction. The resulting stiffness estimates in \Cref{fig:phys-stiffness} capture the expected relationship between compliance and deformation required to reach the target force. The ripe avocado falls into Level 2, while the two unripe avocados fall into Levels 4 and 5, with substantially greater stiffness and resistance encountered over a much smaller jaw displacement. The measurements of the ripe avocado do not overlap with those of the unripe avocados, showing that a single squeeze is sufficient to distinguish them. The two unripe avocados exhibit overlapping measurements, which is consistent with their similar physical stiffness despite differences in visual appearance.

\subsection{API Specification Details}
\label{sec:app_api_phyx}
We provide the complete API specifications for the PhysX modules, including function outputs, documentation strings, and usage examples.

\begin{lstlisting}
    ...
    def get_mass(self, object_name: str) -> float:
        """Measure an object's mass, in kilograms.
    
        Grasps the object, lifts it, infers mass from joint load, averages a
        few readings, and returns the object to where it found it. Gripper
        must be empty before calling.
        Make sure nothing is in the gripper before calling this.
    
        Args:
            object_name: The object to weigh. Molmo2 locates it from this name
                alone, so make it unambiguous - give colour and position
                (e.g. "the left-most red cube"), not just "cube".
    
        Returns:
            Mass in kilograms, +/-30 g noise. Slightly-negative readings on
            very light objects are noise, not failure. A large negative value
            means the grasp failed; nan means no reading was taken - reweigh
            in either case.
    
        Example:
            mass = get_mass("the left-most red cube")
            print(f"left-most red cube: {mass:.3f} kg")
        """
        try:
            mass = self._env.measure_object_mass(object_name)
            print(f"[get_mass] '{object_name}' has mass {mass:.4f} kg")
            save_property_to_knowledge(self._env, object_name, "mass", mass, "kg")
            _log_measurement(self._env, "mass", "_gt_mass", object_name, mass, "kg")
            return mass
        except Exception as exc:
            print(f"Warning: Failed to measure mass of '{object_name}': {exc}")
            return float("nan")

    def get_stiffness(self, object_name: str) -> int:
        """Measure the stiffness of an object by probing its surface.

        This function executes a controlled probing motion on the object,
        measures force and displacement, and classifies the stiffness level.

        Make sure nothing is in gripper before calling this function.

        Returns a stiffness level from 1 (soft) to 5 (rigid):
            1 = Ultra-Soft   (e.g., sponges, foams, soft plush)
            2 = Soft         (e.g., soft rubbers, ripe fruit, silicone)
            3 = Semi-Rigid   (e.g., cardboard, ripe fruit)
            4 = Stiff        (e.g., hardwood, dense polymers)
            5 = Rigid        (e.g., metal, ceramics, stone)
        Returns 0 if measurement fails. You may need to test is again.

        Args:
            object_name: Name of the object to measure (e.g., "red_cube", "apple")
            This function will use Molmo2 for object detecting, it will only take in the current object name as input.
            Make sure that the given object name is clear and will not cause any ambiguity.
            It is adviced that the positional description (left-most, second from right, etc.), color (and object's name if possible) is provided in the object name.

        Returns:
            stiffness_level: Integer 1-5 indicating stiffness. Returns 0 if measurement fails.

        Example:
            stiffness = get_stiffness("apple")
            if stiffness == 2:
                print("Apple stiffness is level 2:soft - likely ripe")
            elif stiffness == 3:
                print("Apple stiffness is level 3:firm - handle carefully")
            elif stiffness == 0:
                print("Apple stiffness is level 0:measurement failed")
        """
        try:
            stiffness = self._env.measure_object_stiffness(object_name)
            print(f"[get_stiffness] '{object_name}' stiffness level: {stiffness}/5")
            _save_property_to_knowledge(self._env, object_name, "stiffness", stiffness, "level")
            return stiffness
        except Exception as e:
            print(f"Warning: Failed to measure stiffness of '{object_name}': {e}")
            return 0
    ...
\end{lstlisting}

\section{Low-Level Perception and Control Primitive Details}
\label{sec:app_control}

\subsection{Accessible Tools}
The coding agent has access to the following low-level APIs:
\begin{itemize}
    \item \texttt{get\_object\_pose(obj\_name)}: Given an object name as a prompt, the current 2D scene observation is passed to Molmo~2, which generates a 2D semantic keypoint. By back-projecting this pixel using the ZED~2i depth map, the function returns a precise 3D spatial pose with a predefined pre-grasp offset.
    \item \texttt{open\_gripper()}: Opens the robotic gripper.
    \item \texttt{close\_gripper()}: Closes the robotic gripper.
    \item \texttt{goto\_pose(pose)}: Given a target end-effector position, the system computes the inverse kinematics (IK) and executes the motion trajectory. 
    \item \texttt{home\_pose()}: Returns the robot arm to its default home configuration.
    \item \texttt{breakpoint\_code\_block()}: Terminates the execution of the current code block, signaling the end of the current reasoning step.
\end{itemize}
The coding agent also has full access to all functions within the standard Python library.

\subsection{API Specifications}
\label{sec:app_api_control}
This section details the complete API specifications for the low-level perception and control primitives, including function signatures, expected outputs, documentation strings, and concrete usage examples.

\begin{lstlisting}
    ...
    def get_object_pose(self, object_name: str) -> tuple[np.ndarray, np.ndarray]:
        """Sample a grasp pose for an object.
        This function will use Molmo2 for object detection; it will only take in the current object name as input.
        Make sure that the given object name is clear and will not cause any ambiguity.
        It is advised that positional descriptions (left-most, second from right, etc.), color, and the object's name (if possible) are provided in the object name.
        Returns:
            position: (3,) XYZ in meters.
            quaternion_wxyz: (4,) WXYZ unit quaternion (often unused for 3DOF setups).
        """
        pos, _ = self._env._get_object_pose(object_name)
        return pos, np.array([1, 0, 0, 0])

    def goto_pose(
        self, position: np.ndarray, quaternion_wxyz: np.ndarray = None, z_approach: float = 0.0
    ) -> None:
        """Go to pose using Cartesian IK provided natively by the AgileX firmware.
        There is no need to call a second goto_pose with the same position and quaternion_wxyz after calling it with z_approach.
        Example:
        goto_pose(np.array([0.1, 0.2, 0.3])) # This controls the arm directly to position [0.1, 0.2, 0.3]
        goto_pose(np.array([0.1, 0.2, 0.3]), z_approach=0.05) # This controls the arm to position [0.1, 0.2, 0.3] + [0, 0, 0.05] and then moves to position [0.1, 0.2, 0.3]
        Args:
            position: (3,) XYZ in meters.
            quaternion_wxyz: (4,) WXYZ unit quaternion. Ignored in 3 DOF positioning.
            z_approach: (float) Z-axis distance offset for the goto_pose insertion approach motion. Will first arrive at position + z_approach meters in the Z-axis before moving to the requested pose. Useful for more precise grasp approaches. Default is 0.0.
        """
        pos = np.asarray(position, dtype=np.float64).reshape(3)
        
        if z_approach != 0.0:
            approach_pos = pos + np.array([0, 0, z_approach])
            self._env.move_to_cartesian_blocking(approach_pos)
            
        self._env.move_to_cartesian_blocking(pos)

    def open_gripper(self) -> None:
        """Open gripper fully."""
        self._env.open_gripper()

    def close_gripper(self) -> None:
        """Close gripper fully."""
        self._env.close_gripper()

    def home_pose(self) -> None:
        """Return the arm to its rest pose."""
        self._env.home_pose()

    def breakpoint_code_block(self) -> None:
        """Call this function to mark a significant checkpoint."""
        return None
    ...
\end{lstlisting}

\section{Planner Agent}
\label{sec:app_planner}

\subsection{Detailed Flow}
Given a task description and a visual observation of the current scene, the Planner Agent first evaluates whether the available information is sufficient to execute the main task and provides a rationale for its decision. If the information is deemed sufficient, the planner directly instructs the coding agent to generate the final execution code. However, if critical physical states remain hidden, the planner generates a JSON-formatted list of task-relevant exploration candidates. For each candidate, the planner specifies the target object, the information to be revealed, and the physical actions required to obtain it. This list of viable candidates is then passed down to the Prioritizer Agent for ranking.

\subsection{Prompt for Planner Agent}
\begin{lstlisting}[
    language={},
    breaklines=true,
    breakatwhitespace=true,
    columns=fullflexible,
    label={lst:planner_prompt}
]
# Your Task
You are a robotics reasoning assistant in an agentic system that helps a robot determine whether it has sufficient information to complete a task based on a single scene image. 
Your output must be a proposed interaction plan. This plan will be passed to a downstream agent that generates control policy code for an AgileX PiPER 6-DOF robot arm.
You will receive:
1. A task description.
2. A scene image.
3. (Optionally) Accumulated scene knowledge and past interaction history.
Your job is to determine whether the robot can complete the task using only the information visible in the image.
If not, please provide instructions on how to interact with or identify the object to obtain more information for completing the task.

# Step Guidance
Follow this reasoning procedure internally:
## Step 1 - Understand the task
Determine the goal of the task and what object(s) are required to complete it.
## Step 2 - Identify task-relevant objects WITH SPATIAL DESCRIPTORS
From the scene image, locate and label all objects, include a SPATIAL DESCRIPTOR so the robot can easily identify which object to interact with:
- Use descriptive spatial terms: "left-most cup", "back-right cup", etc.
## Step 3 - Determine required properties
For each central object, determine what physical properties or hidden information are required to complete the task. These properties may include:
- object mass
- object stiffness
- whether something is hidden inside another object
## Step 4 - Check information sufficiency
Determine whether what you already know is enough to COMMIT TO THE ANSWER the task asks for.

Apply this test to every measurement you have not yet taken: "could its result change my answer?"
- If NO remaining measurement could change the answer, the information IS sufficient. Set "sufficient" to true - even when some objects remain entirely unmeasured, and even when some properties remain unknown.
- If some remaining measurement COULD change the answer, set "sufficient" to false and propose the exploration actions that would resolve it.

Sufficiency is about the decision, not about coverage. A complete table of every property for every object is NOT the goal. Once an object has been ruled out, measuring anything further about it cannot change the answer and is never justified. Do not treat "this property is still unknown" as a reason to keep exploring unless knowing it could actually change what you decide.
## Step 5 - Propose exploration candidates
List all possible candidates for exploration that may provide new task-related information.
Each exploration towards a physical property for individual objects counts as a distinct exploration candidate.
Examples include:
- measuring an object's mass
- measure the stiffness of an object
- moving an object to reveal hidden items

# Guidelines for exploration actions:
- Actions must be directly related to discovering the missing property.
- Actions should focus only on the most task-relevant objects.

# Output format rules (VERY IMPORTANT):
You MUST output a single valid JSON object and NOTHING ELSE.
The JSON schema must be exactly:
{
  "sufficient": boolean,
  "task_complete": boolean,
  "reason": "string explaining why the task can or cannot be completed",
  "central_objects": [
    {
      "name": "object name",
      "description": "short description of the object",
      "required_properties": ["property1", "property2"]
    }
  ],
  "exploration_candidates": [
    {
      "name": "object name",
      "action": "measure_mass | measure_stiffness | relocate | inspect | other",
      "description": "clear description of the exploration action",
      "parameters": {
      "param_name": "type"
    },
      "expected_info": "what information this action reveals",
      "estimated_cost": "low | medium | high"
    },
    ... (#Please list as many candidates as you can.)
  ]
}

# Important constraints:
- If "sufficient" is false, exploration_candidates must contain the actions needed to reveal the missing information.
- Each exploration candidate must be self-contained and executable now, on its own. Do not make a candidate conditional on another candidate or object (no "only if/after ...", no precondition or execution-order fields): which candidate runs next is decided downstream.
- Set "task_complete" to true only if the accumulated knowledge and the current image show the task's goal has already been achieved (e.g. the required object is already at its destination). Then "sufficient" is true, exploration_candidates is empty, and nothing further is executed. If any required action has not been done yet, "task_complete" is false.
- "name" identifies the OBJECT and "action" identifies WHAT IS DONE to it. One candidate per (object, action) pair: measuring an object's mass and measuring its stiffness are two separate candidates that share a "name" and differ in "action".
- Use "other" only when no listed action fits, and make "description" specific enough to tell two "other" candidates on the same object apart.
- Do not output explanations outside the JSON.
- Do not include markdown formatting.
- Do not include additional text before or after the JSON.

\end{lstlisting}

\subsection{Prompt for the Merged Planner and Prioritizer Agent}
As introduced in \Cref{sec:method}, we implemented the \textbf{PhysCaP-joint} baseline for our method ablation studies. For this variant, we merged the standard planner instructions with the prompts from the Prioritizer (detailed in \Cref{sec:app_prioritizer}), applying minor modifications, such as omitting the requirement to generate explicit priority scores.
\begin{lstlisting}[
    language={},
    breaklines=true,
    breakatwhitespace=true,
    columns=fullflexible,
]
# Your Task
You are a robotics reasoning assistant that helps a robot determine whether it has sufficient information to complete a task, and if not, generates a prioritized exploration plan in a single pass.
Your proposed interaction plan will be taken as input for downstream agent to generate robot control code policy.
You are controlling an AgileX PiPER 6-DOF robot arm.
You will receive:
1. A task description.
2. A scene image.
3. (Optionally) Accumulated scene knowledge and past interaction history.
Your job is to:
- Determine whether the robot can complete the task using available information.
- If not, generate exploration candidates already sorted in priority order, so the first candidate in the list is the most efficient action to take next.
- Your goal is to help the robot identify the minimal exploration needed before executing the task, generated in the most efficient execution order.

# Step Guidance
Follow this reasoning procedure internally:
## Step 1 - Understand the task
Determine the goal of the task and what object(s) are required to complete it.
## Step 2 - Identify task-relevant objects WITH SPATIAL DESCRIPTORS
From the scene image, locate and label all objects, include a SPATIAL DESCRIPTOR so the robot can easily identify which object to interact with:
- Use descriptive spatial terms: "left-most cup", "back-right cup", etc.
## Step 3 - Determine required properties
For each central object, determine what physical properties or hidden information are required to complete the task. These properties may include:
- object mass
- object stiffness
- whether something is hidden inside another object
## Step 4 - Check information sufficiency
Determine whether the visible information in the image and the accumulated information obtained so far are sufficient to complete the task.
If NOT sufficient: set "sufficient" to false and proceed to Step 5.
## Step 5 - Generate a PRIORITIZED exploration plan
List all exploration candidates in PRIORITY ORDER (highest priority first). The first candidate in the list will be executed next - make it the single most efficient action available. Apply the following rules while generating:
### Visual Cues First
Exploit visual cues to form hypotheses before committing to physical measurements.
- Visual cues being size, shape, status or any other details related to task descriptions.
- Candidates with cues that relates with the task description most should be prioritized.
- For exmaple small objects will less likely to contain items than bigger objects.
### Physical Property Measurements - Skip Already-Measured Objects
- Do NOT propose or re-execute a measurement (mass, stiffness) on an object that already appears in the accumulated scene knowledge.
- After each measurement, compare the result against already-known values to draw a conclusion (e.g., "lightest cup = empty").
Each exploration towards a physical property for an individual object counts as a distinct candidate.
Examples of exploration actions:
- measuring an object's mass
- measure the stiffness of an object
- moving an object to reveal hidden items

# Guidelines for exploration actions:
- Actions must be directly related to discovering the missing property.
- Actions should focus only on the most task-relevant objects.
- Do not propose unnecessary exploration.
- Prefer the smallest number of actions that would reveal the required information.

# Output format rules (VERY IMPORTANT):
You MUST output a single valid JSON object and NOTHING ELSE.
The JSON schema must be exactly:
{
  "sufficient": boolean,
  "task_complete": boolean,
  "reason": "string explaining why the task can or cannot be completed",
  "central_objects": [
    {
      "name": "object name",
      "description": "short description of the object",
      "required_properties": ["property1", "property2"]
    }
  ],
  "exploration_candidates": [
    {
      "name": "object name",
      "action": "measure_mass | measure_stiffness | relocate | inspect | other",
      "description": "clear description of the exploration action",
      "parameters": {
        "param_name": "type"
      },
      "expected_info": "what information this action reveals",
      "estimated_cost": "low | medium | high"
    }
  ]
}

# Important constraints:
- Each exploration candidate must be self-contained and executable now, on its own. Do not make a candidate conditional on another candidate or object (no "only if/after ...", no precondition or execution-order fields): which candidate runs next is decided by the order of the list.
- Set "task_complete" to true only if the accumulated knowledge and the current image show the task's goal has already been achieved (e.g. the required object is already at its destination). Then "sufficient" is true, exploration_candidates is empty, and nothing further is executed. If any required action has not been done yet, "task_complete" is false.
- "name" identifies the OBJECT and "action" identifies WHAT IS DONE to it. One candidate per (object, action) pair: measuring an object's mass and measuring its stiffness are two separate candidates that share a "name" and differ in "action".
- Use "other" only when no listed action fits, and make "description" specific enough to tell two "other" candidates on the same object apart.
- If "sufficient" is false, exploration_candidates MUST be in priority order (highest priority first).
- The first candidate is the next action to execute .
- Do not output explanations outside the JSON.
- Do not include markdown formatting or code fences.
- Do not include additional text before or after the JSON.
\end{lstlisting}

\subsection{Example Exploration Candidate Lists}
\begin{lstlisting}[
    language={},
    breaklines=true,
    breakatwhitespace=true,
    columns=fullflexible,
]
[
    {
        "name": "can_0",
        "action": "measure_mass",
        "description": "Measure the mass of the blue can to check if it is empty.",
        "parameters": {
            "object_name": "can_0"
        },
        "expected_info": "Mass of the blue can in kg.",
        "estimated_cost": "medium"
    },
    {
        "name": "can_1",
        "action": "measure_mass",
        "description": "Measure the mass of the black can to check if it is empty.",
        "parameters": {
            "object_name": "can_1"
        },
        "expected_info": "Mass of the black can in kg.",
        "estimated_cost": "medium"
    },
    {
        "name": "can_2",
        "action": "measure_mass",
        "description": "Measure the mass of the green can to check if it is empty.",
        "parameters": {
            "object_name": "can_2"
        },
        "expected_info": "Mass of the green can in kg.",
        "estimated_cost": "medium"
    },
    {
        "name": "can_3",
        "action": "measure_mass",
        "description": "Measure the mass of the red can to check if it is empty.",
        "parameters": {
            "object_name": "can_3"
        },
        "expected_info": "Mass of the red can in kg.",
        "estimated_cost": "medium"
    }

]
\end{lstlisting}

\section{Prioritizer Agent}
\label{sec:app_prioritizer}
\subsection{Detailed Flow}
Given a list of exploration candidates, the Prioritizer Agent evaluates and ranks them using visual heuristics. It assigns each candidate a priority score based on visual plausibility and relevance to the main task, alongside a textual justification for the score. The candidate list is then sorted in descending order by this priority score. The highest-priority candidate is selected for immediate exploration and dispatched to the Coding Agent as a subtask.

\subsection{Prompt for Prioritizer Agent}
\begin{lstlisting}[
    language={},
    breaklines=true,
    breakatwhitespace=true,
    columns=fullflexible,
    label={lst:prioritizer_prompt}
]
# Your Role: 
You are an expert robotics reasoning assistant. Your job is to reorder a list of candidate interaction steps (provided as JSON) so that the robot acquires the most decision-relevant information with the least interactions.
You are controlling an AgileX PiPER 6-DOF robot arm interacting with objects on a table.

# Hard Constraints (apply before ranking)
## Visual Cues First
Exploit visual cues to form hypotheses before committing to physical measurements.
- Visual cues being size, shape, status or any other details related to task descriptions.
- Candidates with cues that relates with the task description most should be prioritized.
- For exmaple small objects will less likely to contain items than bigger objects. 
## Physical Property Measurements - Skip Already-Measured Objects
- Do NOT propose or re-execute a measurement (mass, stiffness) on an object that already appears in the accumulated scene knowledge.
- After each measurement, compare the result against already-known values to draw a conclusion (e.g., "lightest cup = empty").


# Ranking Ladder (apply strictly in order)
Rung 1 decides the ordering. Rung 2 is consulted ONLY to break ties within Rung 1.

## Rung 1 - Eliminative Power
Rank a candidate by how much of the remaining uncertainty its result can remove.

- A candidate whose outcome PARTITIONS the remaining candidates - separating those
  that can still be the answer from those that cannot - outranks one that only adds
  detail about a single object.

- Do NOT fully characterise one object while other candidate objects remain entirely
  unmeasured. Establish the discriminating property across all candidates first, then
  return for follow-up properties only on the survivors.

- If the accumulated scene knowledge already rules an object out, every remaining
  candidate on that object ranks LAST: its result can no longer change the answer.

## Rung 2 - Execution Cost (tie-break only)
Among candidates of EQUAL eliminative power, prefer the cheaper action. Cost means
ROBOT EXECUTION TIME on the physical arm. Ignore the time taken by your own
reasoning - it is not part of the task cost.

Approximate robot execution time per action (AgileX PiPER):
  measure_mass     ~15 s   grasp, read torque, replace
  measure_stiffness  ~40 s   home, fine contact detection, then probing
  relocate           ~10 s   pick and place
  inspect            ~5 s    move and observe

Travel distance and kinematic convenience are NOT ranking criteria. Do NOT prefer a
candidate merely because the arm is already near that object.


# Output Rules
- Return ONLY a valid JSON list of the candidate steps in the new priority order.
- Do NOT modify the name, action, description, parameters, estimated_cost, or expected_info fields of any step.
- `name` is the object and `action` is what is done to it, so two steps may share a `name` and differ only in `action`. Always echo BOTH fields back: a step is identified by the pair, and a name alone is ambiguous.
- Do NOT add or remove steps - only reorder them.
- If two candidates have equal priority, preserve their relative original order.

# Explain-First Mode
First consider each candidate and explain why it should be prioritized. Then return a single JSON object with:
- prioritized_candidates: the reordered list
- reasons: an array of {name, reason} entries explaining the ordering
When ordering the candidates, the first candidate will have the hightest priority, meaning that this candidate will be executed next.
Do NOT change candidate fields, add/remove candidates, or output any extra text.

\end{lstlisting}

\subsection{Example Prioritized Candidate Lists}
\begin{lstlisting}[
    language={},
    breaklines=true,
    breakatwhitespace=true,
    columns=fullflexible,
]
{
  "scored_candidates": [
    {
      "name": "can_2",
      "action": "measure_mass",
      "reason": "The green can has a straw, indicating it is open. Open cans are much more likely to be empty than unopened ones, so it should be prioritized for mass measurement.",
      "priority_score": 10
    },
    {
      "name": "can_3",
      "action": "measure_mass",
      "reason": "The red can also has a straw, indicating it is open. It is highly likely to be empty compared to the unopened cans.",
      "priority_score": 9
    },
    {
      "name": "can_0",
      "action": "measure_mass",
      "reason": "The blue can appears unopened, making it very unlikely to be the empty can. It should only be measured if the open cans are not empty.",
      "priority_score": 2
    },
    {
      "name": "can_1",
      "action": "measure_mass",
      "reason": "The black can appears unopened, making it very unlikely to be the empty can. It has the lowest priority.",
      "priority_score": 1
    }
  ]
}
\end{lstlisting}

\section{Coding Agent}

\subsection{Generating Code Policies}
For code policy generation, we use CaP-Agent0 \cite{fu2026cap} as our base method. Specifically, we employ the M3 setting, but provided APIs at a high-level abstraction (similar to ``S2''). ``M3'' denotes the use of a Visual Differencing Module to interpret scene changes between turns rather than directly processing 2D image inputs. The coding agent receives documentation strings (detailed in \Cref{sec:app_api_phyx} and \Cref{sec:app_api_control}), task-specific prompts (\Cref{sec:app_task_prompts}), and general prompts (\Cref{sec:app_cap_prompts}).

\subsection{Coding Agent General Prompts}
\label{sec:app_cap_prompts}
When generating a code policy for an exploration candidate, the coding agent receives a prompt containing the candidate's context, recently obtained physical properties, and the main task objective.
\begin{lstlisting}[
    language={},
    breaklines=true,
    breakatwhitespace=true,
    columns=fullflexible,
]
Selected next high-level step from the reasoning pipeline:
- name (the object to act on): {candidate_name}
- action (what to do to it): {candidate_action}
- description: {candidate_desc}
- expected_info: {candidate_info}
- estimated_cost: {candidate_cost}
- parameters: {candidate_params}

Original task:
{original_task}

Your job is to generate Python code for ONLY the selected next high-level step.
Perform exactly this one action on exactly this one object. Do not measure any
additional property, and do not act on any other object, even when doing so in
the same code block would be more efficient - the reasoning pipeline schedules
those as their own steps, and doing them here corrupts its accounting.
Do not plan the whole task from scratch unless the selected candidate is explicitly the original task.
Use the available APIs from the prompt above.
The code should be executable Python only.
\end{lstlisting}

If the planner determines that the current information is sufficient to complete the task and no further exploration is necessary, the following prompt is issued instead:
\begin{lstlisting}[
    language={},
    breaklines=true,
    breakatwhitespace=true,
    columns=fullflexible,
]
The reasoning pipeline has determined that sufficient information has been gathered
to complete the FULL task without additional exploration.

=== ACCUMULATED SCENE KNOWLEDGE ===
{scene_knowledge}
=== END SCENE KNOWLEDGE ===

Original task:
{original_task}

Your job is to generate Python code that completes the ENTIRE task from start to finish
using the accumulated knowledge above. Do not perform unnecessary exploration steps -
you already know what you need. Use the available APIs from the prompt above.
The code should be executable Python only.
\end{lstlisting}

Additionally, an execution guidance prompt is provided in both scenarios to ensure reliable task completion.
\begin{lstlisting}[
    language={},
    breaklines=true,
    breakatwhitespace=true,
    columns=fullflexible,
]
# Key rules:
- Write code to interact with the environment. 
- You will also receive visual feedback of the table. Examine the image.
- IMPORTANT: To remember the value for your next step, you MUST explicitly `print()` the result so you can read it in the console stdout.

- Call get_object_pose(object_name) to find objects.
- Going back to home pose after putting back the interacted object on the table is suggested after each interaction.
- When moving objects, move 10 cm above the object before going to the target position to avoid collisions.
- When placing object on for the wooden tray, aim for the object's z-coordinate + 5 cm above the tray's surface to avoid collisions.
- After placing objects on the wooden tray, lift arm 20 cm above the tray and object before going to home pose to avoid collisions.
- Go to home position after you've successfully completed the task.
\end{lstlisting}

\section{Common Failure Modes and Implementation Limitations}
\label{sec:failure_modes}

\subsection{Perception and Execution}
Our control pipeline relies on language-conditioned object localization, so an underspecified description from the coding agent becomes ambiguous when several objects look alike, as in the avocado task, causing the wrong object to be measured. Imperfect grasping can likewise abort execution or corrupt the resulting measurement. Repeated measurements within each PhysX call improve robustness to mechanical noise but do not eliminate such errors, which then propagate through the system belief and lead to incorrect decisions.
These failures are compounded by limitations in the current perception and control stack. Object localization depends on 2D predictions \cite{clark2026molmo2} mapped to single-camera depth, so 2D errors directly degrade 3D end-effector precision. Future iterations will incorporate multi-view or 3D-native models. Hardware communication latency with the PiPER arm occasionally causes physical trajectories to diverge from generated code, highlighting the need for real-time control signal management.

\subsection{Agentic Reasoning}
The Prioritizer sometimes fails to exploit useful visual cues and reverts toward exhaustive exploration. Although explicit belief updates improve measurement tracking, maintaining consistent object-property associations becomes harder over longer task horizons. More generally, reliance on commercial VLM APIs \cite{team2024gemini} introduces unpredictable latency and reasoning variability. Transitioning to locally hosted open-source models \cite{bai2025qwen3} would ensure operational consistency.

\end{document}